\documentclass{article}
\PassOptionsToPackage{numbers,compress}{natbib}
\usepackage[preprint]{neurips_2026}

\providecommand{\tightlist}{\setlength{\itemsep}{0pt}\setlength{\parskip}{0pt}}
\newcommand{\pmci}[2]{#1{\scriptscriptstyle\textcolor{gray!70}{(\pm #2)}}}
\newcommand{\pmcibest}[2]{\textbf{#1}{\scriptscriptstyle\textcolor{gray!70}{(\pm #2)}}}
\newcommand{\best}[1]{\textbf{#1}}

\usepackage{amsmath,amssymb}
\usepackage{graphicx}
\usepackage{booktabs}
\usepackage{url}
\usepackage{hyperref}
\usepackage{xcolor}         % colors

\usepackage{algorithm}
\usepackage{algpseudocode} % algorithmic
\usepackage{array}

\usepackage{enumitem}

\title{Completion at the Boundary (CaB): Deployable Switching with Completion-Aware Control under Limited Calibration}
\author{
Yusuke Sano, Takeshi Itoga \\
Intelligent Systems Laboratory, SECOM Co., Ltd. \\
\texttt{yus-sano@secom.co.jp, t-itoga@secom.co.jp}
}

\begin{document}
\maketitle

\begin{abstract}
Vision--language--action (VLA) agents can execute natural-language
instructions, yet deployed systems still lack an operational interface:
deciding when the instruction is complete. This gap is acute in short
composites (``do A, then B''), where mistimed handoffs cascade into
downstream failures. Completion is inherently closed-loop because
switching is an intervention that changes the instruction context and
thus future actions and observations. We study completion under a
deployable low-calibration regime motivated by open-ended instruction
spaces, enforcing no test-time relearning and a single globally
calibrated switching rule selected once on development set and reused
unchanged on test set. Under this constraint, collapsing asymmetric
boundary evidence into a single scalar can be brittle under polarity
shifts across tasks. We propose \textbf{Completion at the Boundary
(CaB)}, which predicts an event-local completion object in the form of
\textbf{Boundary-Phase Tokens} (Before/Hit/After), retaining two-sided
boundary evidence under this discipline. \textbf{CaB-When} converts this
completion object into a minimal, auditable switching decision
(\emph{when}), while \textbf{CaB-How} reuses the same completion object
to condition action generation for boundary-stable control through
handoffs (\emph{how}). Using an intervention-aware \textbf{E1/E2}
protocol, we show that CaB improves composite execution and handoff
quality on a first-person Minecraft VLA benchmark under matched capacity
and deployability constraints.
 % 0.5 page
\end{abstract}

\section{Introduction}\label{sec:intro}

\begin{figure*}[t]
  \centering
  \includegraphics[width=\textwidth]{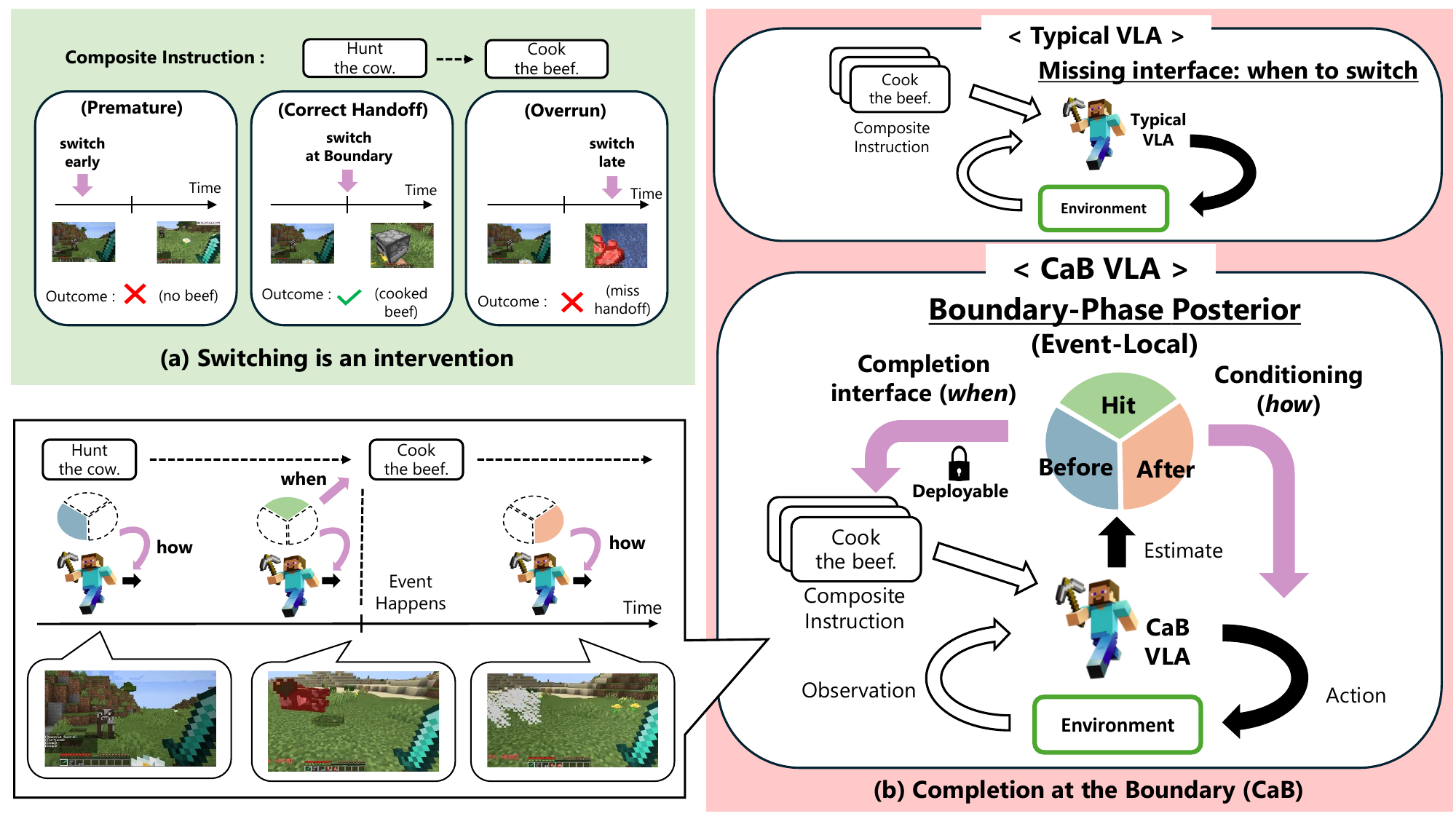}
\caption{\textbf{Key idea: deployable completion interface under intervention.}
(\textbf{a}) In short composite instructions (“do A, then B”), mistimed handoffs (premature vs.\ overrun) cause downstream failures; switching updates the active instruction and thus intervenes in the trajectory.
(\textbf{b}) Many VLA policies lack an explicit completion interface for deciding \emph{when} to switch, whereas \textbf{CaB} learns an event-local \textbf{boundary-phase posterior} (Before/Hit/After) and uses it for both a deployable completion interface (\emph{when}) and control conditioning for boundary-stable execution (\emph{how}).}
\label{fig:keyidea}
\end{figure*}

Vision--language--action (VLA) policies have rapidly advanced in
executing natural-language instructions in embodied environments,
ranging from end-to-end robot control to open-ended long-horizon
instruction following with hierarchical VLA systems
\citep{RT1,RT2,OpenVLA,pi,SayCan,HiRobot}. Similar progress is emerging
in open-world Minecraft, where recent VLA agents can follow diverse
human instructions across large collections of atomic tasks
\citep{JarvisVLA,STEVE1,Optimus2,OpenHA}.

Yet deployed systems still lack a basic operational interface: deciding
when the current instruction is complete. This gap is acute in short
composite routines (``do A, then B''), where a mistimed handoff can
cascade into downstream failures.

A key reason completion is challenging in composites is that switching
is an intervention: changing the active instruction alters the agent's
context, actions, and thus future observations. Completion is therefore
inherently closed-loop, with completion and control tightly coupled,
rather than a passive prediction problem. In composites, completion
decisions directly change the active instruction, making switching a
high-impact intervention that can materially affect downstream outcomes.
For reliability and safety \citep{Classifier-SAFE}, completion should
therefore be exposed as a \textbf{deployable completion interface} for
intervention-sensitive execution---not merely as a passive prediction
head.

We target a deployability regime motivated by practical constraints in
open-ended instruction spaces, where new phrasings, compositions, and
user-specific goals continually emerge. Task/group-specific calibration
would therefore require ongoing maintenance of the task/group taxonomy
and calibration mappings, making it operationally ill-defined; moreover,
introducing explicit grouping (manual or learned) increases calibration
capacity and undermines auditability. We therefore enforce a strict
low-capacity \textbf{deployability discipline}---no test-time relearning
and switching via a single global rule \((\theta,L)\) calibrated once on
development (dev) set and reused unchanged on test set
(Sec.\ref{sec:discipline}).

Under this discipline, a single global rule becomes brittle when
completion evidence is collapsed into a scalar/bit in a way that
discards boundary-phase structure, trading premature/false switches
against overruns (e.g.,
\citep{STG-SELF,STG-RL,STG-VLM,Classifier-SAFE,Classifier-VLM,Classifier-VLA}).
This happens because boundary evidence can be polarity-shifted across
tasks---some provide mainly anticipatory evidence before first success,
others mainly confirmatory evidence after. To handle polarity shifts, we
therefore retain two-sided boundary evidence in an \emph{event-local
posterior over boundary phase} (Before/Hit/After). In practice, we
realize this posterior as an online posterior over a small, discrete
\textbf{Boundary-Phase Tokens (BPT)} vocabulary. This brittleness claim
is scoped to the matched deployability discipline;
Appendix\textasciitilde{}\ref{sec:polarity} gives simple constructions
underscoring the importance of retaining two-sided boundary evidence.

Crucially, composite execution concentrates at handoff boundaries, where
switching is an intervention: the agent must decide \emph{when} to hand
off, and also remain stable \emph{through} the handoff as the
instruction context changes. A completion signal such as a posterior
over BPT is necessary to make \emph{when} decisions, but by itself it
does not guarantee stable behavior \emph{through} the intervention.
Thus, a completion signal is necessary but not sufficient for
closed-loop composite success.

To meet \emph{both} requirements, we propose \textbf{Completion at the
Boundary (CaB)} (Fig.\textasciitilde{}\ref{fig:keyidea}). CaB trains a
single autoregressive VLA policy that jointly predicts actions and a BPT
posterior as a shared completion object, and \emph{dual-uses} it for
switching and control: \textbf{CaB-When} reads it to decide \emph{when}
to switch, while \textbf{CaB-How} reuses it to condition \emph{how}
actions traverse the handoff. We validate CaB with an intervention-aware
\textbf{E1/E2} protocol that separates action-fixed detection (E1) from
closed-loop execution (E2), and attribute composite gains via a
Switching×Conditioning decomposition.

\textbf{Contributions.}

\begin{itemize}
\item
  \textbf{One completion object: dual-use for switching and control.} We
  introduce a completion object---an online posterior over
  Boundary-Phase Tokens (BPT) that \emph{retains two-sided boundary
  evidence} (Before/Hit/After)---used for a completion interface
  (\emph{when}) and completion-aware control (\emph{how}).
\item
  \textbf{CaB: two consumers of the same completion object.} We propose
  Completion at the Boundary (CaB), which jointly predicts actions and a
  BPT posterior; CaB-When uses it to decide \emph{when} to switch under
  the deployability discipline, while CaB-How reuses it to condition
  \emph{how} actions traverse the handoff for boundary-stable control.
\item
  \textbf{Intervention-aware protocol and empirical validation in
  Minecraft.} We introduce an E1/E2 protocol and a When (Switching) ×
  How (Conditioning) decomposition to separate detection from
  closed-loop effects and attribute gains. we report empirical evidence
  for CaB in a first-person Minecraft VLA setting under matched
  deployability constraints.
\end{itemize}

\section{Related Work}\label{sec:related_work}

\textbf{A unifying view: temporal structure for completion.} Across
VLA/RL, temporal structure relevant to completion appears in three
forms: (i) supervision-based auxiliary targets such as progress or
step-to-event regression that provide dense scalar signals
\citep{Supervision-TDM,STG-SELF,Classifier-SAFE}, (ii) prediction-based
timing models that estimate event times/hazards from observations
\citep{Hazard,NeuralHawkes,TPP}, and (iii) control-internal temporal
abstraction mechanisms (e.g., options) in which termination governs when
to hand off \citep{Options,Option-Critic}. However, none provides what
we target: a deployable completion interface that exposes boundary
evidence under intervention with limited calibration capacity.

\textbf{Sequence modeling in vision--language--action.} Recent VLA
systems treat control as action token prediction conditioned on
multimodal context \citep{RT1,RT2,OpenVLA,pi}; hierarchical stacks
further emphasize long-horizon execution via decomposition and
delegation (e.g., Hi Robot) \citep{SayCan,HiRobot}. In open-world
Minecraft, public VLA models (e.g., Jarvis‑VLA, Optimus2) shows that
large vision--language models can be post-trained to execute diverse
game instructions with keyboard-and-mouse control
\citep{JarvisVLA,STEVE1,OpenHA,Optimus2}. Our work is orthogonal to
scaling tokenization or pretraining: we focus on the missing
\emph{operational interface} for composites---deciding \emph{when} the
current instruction is complete.

\textbf{Supervision-based temporal signals (step-to-event / progress).}
Such supervision-based temporal objectives range from TD--based
horizon-conditioned value prediction to auxiliary progress estimation
learned from interaction \citep{Supervision-TDM,Supervision-Navi}. In
reinforcement learning, such progress/step-to-event heads are often used
as global (or relative \cite{PWD-VLA,PWD-RL}) progress estimates for
reward shaping \citep{STG-SELF,STG-RL,STG-VLM}, rather than as an
explicit completion interface. A coarse alternative is a thresholded
binary done head \citep{Classifier-SAFE,Classifier-VLM,Classifier-VLA}.
CaB instead is designed as an explicit completion interface, predicting
a boundary-phase posterior (Before/Hit/After) that preserves two-sided
evidence.

\textbf{Prediction-based temporal modeling (hazard / point processes).}
Time-to-event models (e.g., Cox-style hazards and neural point
processes) capture uncertainty in event timing and temporal dependencies
\citep{Hazard,NeuralHawkes,TPP}. These methods are typically evaluated
under \emph{passive observation}, where prediction does not influence
the future trajectory. Our completion setting is \emph{closed-loop}:
switching intervenes on the trajectory, so completion must be evaluated
under intervention rather than as passive event prediction.

\textbf{Control-internal temporal abstraction (options and
termination).} Hierarchical RL introduces temporal abstraction via
temporally extended actions and termination: options (and option-critic)
learn both \emph{when} to terminate and \emph{how} behavior evolves
within an option \citep{Options,Option-Critic}. CaB is aligned in
treating phase/termination as central to composition, but differs in
exposure: rather than keeping temporal structure implicit inside the
controller, it exposes a BPT posterior as an explicit completion object
readable by a deployable completion interface.

\textbf{Provided conditioning signals vs.~inferred completion
posteriors.} Prior work often stabilizes staged behavior by conditioning
policies on externally provided signals---e.g., explicit goals
(including goal images) or target returns/reward-to-go---available from
the task interface or dataset
\citep{Cond-Value,Cond-Hindsight,Cond-Image,Cond-Transformer,Cond-OfflineRL}.
Our setting differs in that such variables are not provided: CaB infers
a BPT posterior online and dual-uses it both as a deployable completion
interface (\emph{when}) and as a control-conditioning signal for
boundary-stable handoffs (\emph{how}).

\section{Problem Setting}\label{sec:problem}

We consider discrete-time interaction. At step \(t\), the agent observes
\(o_t\), receives the currently active natural-language instruction
\(l^{(i_t)}\), and outputs an action \(a_t\). Many VLA policies map
\((o_t,l^{(i_t)})\) to actions \(a_t\) but typically lack an explicit
operational interface for deciding when the current instruction is
complete; we add such a deployable completion interface that uses an
online completion signal to update the instruction index \(i_t\).
Because updating \(i_t\) intervenes on the instruction context,
switching changes future observations and actions, so completion is
inherently closed-loop.

\textbf{Scope \& Assumption (latent completion time; offline
annotation).} We focus on tasks where each sub-instruction admits a
completion time \(t_i^*\) defined by the first success event. During
training and evaluation, we assume access to an offline procedure (e.g.,
state predicates or manual labeling) that can assign \(t_i^*\). These
annotations are used only to construct supervision and to evaluate
timing; deployment relies solely on observations and instructions.

\section{Method: CaB --- Dual-Use Boundary-Phase
Posterior}\label{sec:method}

\begin{figure*}[t]
\centering
\includegraphics[width=\textwidth]{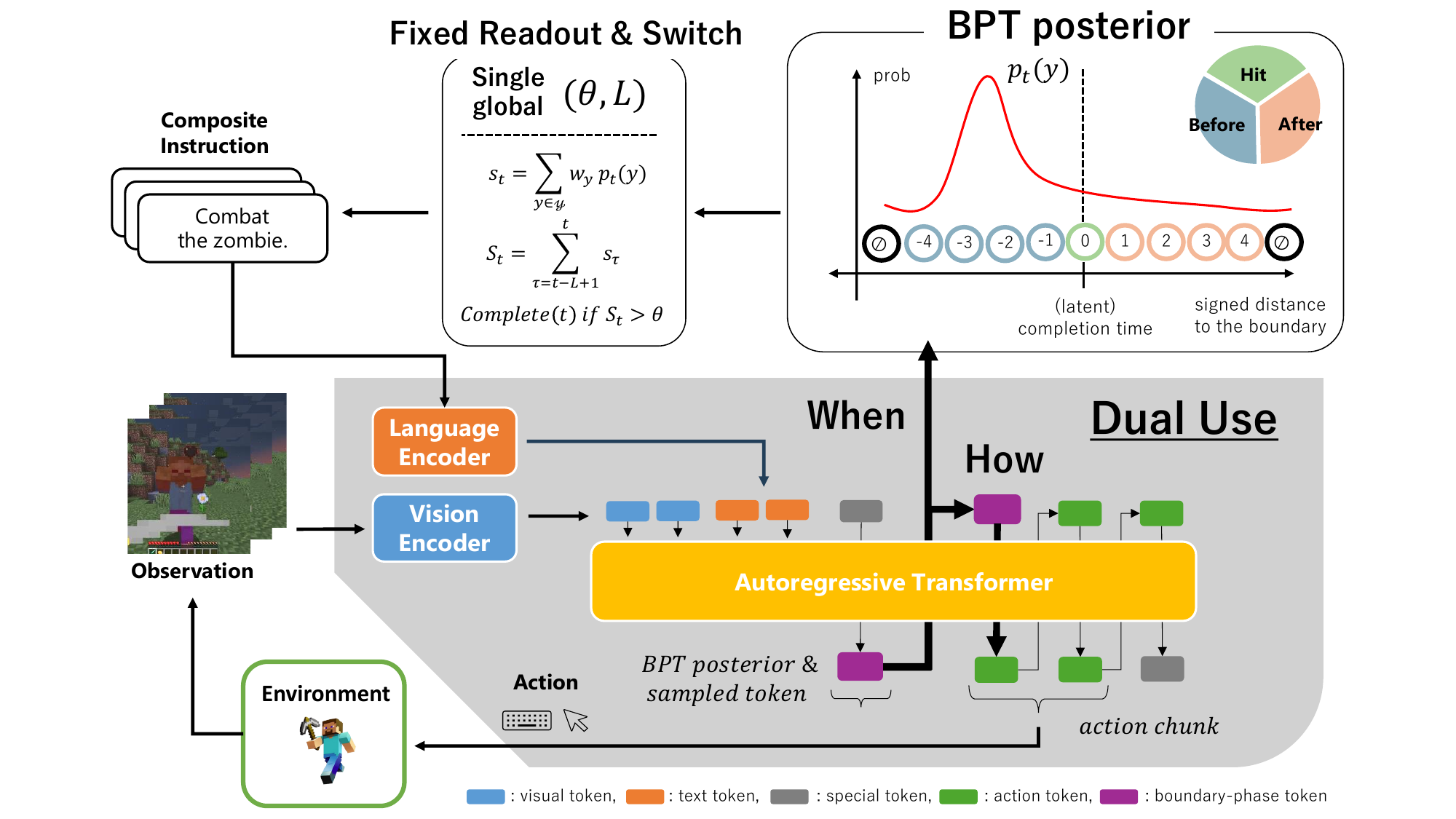}
\caption{\textbf{CaB overview (one posterior, two consumers).}
An autoregressive VLA policy jointly predicts actions and an event-local BPT posterior $p_t(y)$ as the shared completion object.
This posterior is dual-used by two consumers: (i) CaB-When applies a fixed readout to decide \emph{when} to switch, and (ii) CaB-How reuses it to condition \emph{how} actions are generated for boundary-stable control.}
\label{fig:overview}
\end{figure*}

CaB trains a single autoregressive VLA policy (parameterized by
\(\phi\)) that jointly predicts actions and a posterior over
Boundary-Phase Tokens (BPT) anchored to the first-success event, which
serves as the shared completion object. At inference, we dual-use this
same posterior via two consumers: CaB-When exposes it as a deployable
completion interface for switching (\emph{when}) under a fixed
low-capacity rule, while CaB-How reuses it to condition action
generation for boundary-stable control (\emph{how}); see
Fig.\textasciitilde{}\ref{fig:overview} for an overview of CaB.

\subsection{The Completion Object: Boundary-Phase Tokens
(BPT)}\label{sec:bpt}

Because boundary evidence can be polarity-shifted across tasks, we
retain two-sided boundary evidence as an event-local posterior over
boundary phase (Before/Hit/After). We instantiate this posterior with
Boundary-Phase Tokens (BPT), a small discrete vocabulary that bins
boundary phase around the first-success event within a fixed window.
Many progress-style approaches emphasize pre-event evidence
(Before/Hit), while post-event evidence is often left implicit
\citep{STG-SELF,STG-RL,STG-VLM,PALM}.

Under our problem setting, each sub-instruction admits an
offline-annotatable completion time \(t_i^*\) (first success), which is
used only to construct supervision; at inference, \(t_i^*\) is
unobserved and the agent must infer the BPT posterior online from
past-only context.

\textbf{Event anchoring (offline).} Given an annotated completion time
\(t^*\), we define the signed distance \(d_t=t-t^*\) to the boundary and
discretize it into a BPT label \(y_t\in\mathcal{Y}\) within an
event-local window.

Using a window radius \(K\) (default \(K=20\) steps; 1s at 20 Hz) and
fixed bin edges, we assign \(y_t\in\mathcal{Y}\) by: (i)
\(y_t=\mathrm{Hit}\) if \(|d_t|\le 1\); (ii) else if
\(2\le |d_t|\le K-1\), set \(b=\left\lfloor |d_t|/2\right\rfloor\) and
assign \(y_t=\mathrm{Before}[b]\) if \(d_t<0\) or
\(y_t=\mathrm{After}[b]\) if \(d_t>0\); and (iii)
\(y_t=\varnothing_{\mathrm{ow}}\) otherwise (out of window), i.e., if
\(|d_t|\ge K\).

Importantly, BPT uses a single shared vocabulary across tasks; it
encodes only signed proximity to completion, not task-specific
semantics.

\textbf{Posterior as a reusable completion object.} The model predicts a
posterior over BPT classes from past-only context \(c_t\) (observation
\(o_t\), active instruction \(l^{(i_t)}\), and optional history up to
\(t\)): \[
p_t(y) \;\;=\;\; P_\phi(y_t = y \mid c_t), \qquad y\in\mathcal{Y}.
\] This posterior is the single completion object reused by both
consumers: CaB-When (switching) and CaB-How (control conditioning).

\subsection{\texorpdfstring{Consumer A (\emph{when}): CaB-When as a
deployable completion
interface}{Consumer A (when): CaB-When as a deployable completion interface}}\label{sec:cab_when}

CaB-When exposes the posterior as a deployable, auditable switching
interface under limited calibration capacity: no test-time relearning, a
fixed readout, and a single globally calibrated rule \((\theta, L)\)
selected once on dev set and reused unchanged on test set.

\textbf{Fixed linear readout (low capacity, auditable).}~ Let
\(p_t(y)=P_\phi(y_t=y\mid c_t)\) denote the inferred BPT posterior for
the currently active sub-instruction from past-only context \(c_t\). ~
We compute a scalar completion-evidence score via a transparent fixed
linear readout: \[
s_t=\sum_{y\in\mathcal{Y}} w_y\,p_t(y).
\] We use a predetermined triangular kernel over signed distance (low
capacity, auditable). Let \(\delta(y)\in\mathbb{Z}\) denote the
representative signed distance associated with BPT class
\(y\in\mathcal{Y}\) (bin-center distances). We define
\[w_y=\max(0,\,K-|\delta(y)|),\] with the same event-local radius \(K\)
used for BPT windowing.

This fixed kernel can be viewed as a distribution-level proximity
score---i.e., it measures posterior mass near the boundary rather than
the proximity of a point estimate (see
Appendix\textasciitilde{}\ref{sec:proximity-of-mean}). We also report
one-sided (Before/After-only) kernels and a learnable readout as
ablations in Sec.\ref{sec:ablation}.

\textbf{Learning-free aggregation and global thresholding.} To reduce
sensitivity to spikes without learning, we aggregate over a short
horizon \(L\):
\[S_t=\sum_{\tau=t-L+1}^{t} s_\tau,\qquad \text{Complete}(t)\iff S_t \ge \theta.\]
We calibrate the single global pair \((\theta, L)\) once on dev set and
reuse it unchanged on test set. ~

\textbf{Minimal switch wrapper.} We use a deterministic one-shot update:
if \(\mathrm{Complete}(t)\) fires for the current sub-instruction, we
advance \(i_t\) by exactly one (capped at the final sub-instruction);
otherwise we keep \(i_t\) unchanged. By default we use zero hold
(\(H=0\)); optionally, we insert a fixed hold of \(H\) steps
(\(0\le H < K\)) after a trigger for smoother CaB-How handoff.

\subsection{\texorpdfstring{Consumer B (\emph{how}): CaB-How as
completion-aware control
conditioning}{Consumer B (how): CaB-How as completion-aware control conditioning}}\label{sec:cab_how}

Completion prediction alone is not sufficient for composite success: at
the handoff boundary, the agent must also remain stable while the
instruction context changes. CaB-How addresses this by reusing the same
inferred completion object for control-side conditioning, improving
boundary-stable behavior in closed loop.

\textbf{BPT--Action factorization: predict BPT first, then act.} At each
step \(t\), CaB-How models two within-step tokens---a BPT token \(y_t\)
and an action token \(a_t\)---given past-only context \(c_t\), and
factorizes \[
p_\phi(y_t, a_t \mid c_t)=p_\phi(y_t \mid c_t)\,p_\phi(a_t \mid c_t, y_t).
\] The BPT prediction term \(p_\phi(y_t\mid c_t)\) defines the shared
completion object as a posterior \(p_t(\cdot)\), which CaB-When reads
for switching; CaB-How reuses it by conditioning action generation on a
BPT token via \(p_\phi(a_t\mid c_t,y_t)\).

\textbf{Within-step one-way masking (no reverse leakage).} We implement
CaB-How in a single autoregressive transformer with one-way within-step
attention: at step \(t\), actions may attend to same-step phase
(phase\(\rightarrow\)action), but phase cannot attend to same-step
actions (no action\(\rightarrow\)phase). Thus reverse leakage is
structurally impossible: the completion posterior used by CaB-When is
computed before \(a_t\) and cannot depend on the action.

\textbf{How-ON/OFF.} How-ON conditions action generation on \(y_t\);
How-OFF keeps the same BPT prediction and loss but drops \(y_t\) from
the action conditioning, isolating the effect of control-side reuse.
Many prior VLA/RL designs predict phase and actions with separate heads
(or even separate modules); How-OFF matches this decoupled setting
\citep{STG-SELF,PALM,pi_six}.

\subsection{Training and Inference: one posterior, two
consumers}\label{training-and-inference-one-posterior-two-consumers}

\textbf{Training (shared across consumers).}

We train a single autoregressive model that predicts both the BPT token
and the action token at each step. Because BPT is a discrete token, it
can be learned \emph{without architectural friction} in
transformer-style sequence modeling. Under teacher forcing, the loss is
the sum of next-token cross entropies: \[
\mathcal{L} = \sum_t \Big(
\mathrm{CE}(y_t,\; p_\phi(y_t\mid c_t))
\;+\;
\mathrm{CE}(a_t,\; p_\phi(a_t\mid c_t, y_t^{\mathrm{cond}}))
\Big),
\] where \(y_t\) is the event-derived BPT label and
\(y_t^{\mathrm{cond}}\) is the token fed to the action head (typically
the ground-truth \(y_t\)). The first term trains BPT-token prediction
and yields the BPT posterior \(p_t(\cdot)=p_\phi(\cdot\mid c_t)\), read
out by CaB-When for switching and reused by CaB-How for control
conditioning. The second term trains the same model to predict the
action token conditioned on \(y_t^{\mathrm{cond}}\).

\textbf{Inference (one posterior, two consumers).}

At each step \(t\), a single autoregressive VLA policy \(p_\phi\) first
produces the BPT posterior \(p_t(\cdot)=p_\phi(\cdot\mid c_t)\);
CaB-When reads out \(p_t(\cdot)\) with the fixed wrapper to decide
switching, while CaB-How uses the same policy to sample a BPT token from
\(p_t(\cdot)\) and generates an action conditioned on that token.

\section{Evaluation Protocol}\label{sec:evaluation}

We use an intervention-aware E1/E2 protocol to separate
completion-signal quality from closed-loop execution effects under
matched deployability constraints. See
Appendix\textasciitilde{}\ref{sec:evaluation_detail} for details.

\subsection{Evaluation: Separating Detection from Closed-Loop
Effects}\label{sec:e12}

Because switching is an intervention (Fig.\ref{fig:keyidea}(a)), we
evaluate completion via two complementary views: \textbf{E1 (necessary;
action-fixed detection)} and \textbf{E2 (sufficient; closed-loop
execution)}. In E1, we evaluate completion-signal quality on a shared
bank of fixed trajectories---a common set of rollouts (observations and
sampled actions) reused across all methods---thereby isolating
prediction quality from behavioral feedback and ensuring matched
evaluation conditions. In E2, we run the agent with switching enabled
and measure end-to-end closed-loop execution outcomes and boundary
timing effects under intervention.

\textbf{Metrics and diagnostics.} In E1, we report Completion-F1 within
a fixed \(\pm 20\)-step tolerance window around ground truth completion
time \(t^*\) and False Completion (episode-level). In E2, we report
Single and Composite Task Success Rate, with timing errors
(episode-level): we count a Premature episode if
\(t_{\mathrm{switch}} < t^* - 20\) and an Overrun episode if
\(t_{\mathrm{switch}} > t^* + 20\), and report Premature/Overrun rates
over episodes. We also report handoff quality
\(\mathrm{SR}_{2\mid 1}=\Pr(\text{subtask 2 succeeds}\mid \text{subtask 1 succeeds})\).

\subsection{Deployability discipline}\label{sec:discipline}

\textbf{Core discipline.} All methods are evaluated under the same
deployability constraints: (i) no test-time relearning, (ii) a fixed,
low-capacity switching wrapper that converts each method's completion
signal into switching decisions, and (iii) a single global switching
rule \((\theta, L)\), calibrated once on dev set and reused unchanged on
test set.

\textbf{Calibration protocol.} We select \((\theta, L)\) on dev set by
maximizing E1:Completion-F1 (rather than E2) and freeze it for all test
evaluations.

\textbf{Why a single global \((\theta,L)\) with E1.} In open-ended
instruction spaces, task/group-specific calibration is operationally
ill-defined (it would require ongoing maintenance), and explicit
grouping increases calibration capacity and undermines auditability. We
therefore use a single global \((\theta,L)\) to minimize calibration
capacity while keeping the completion interface stable and auditable.
Calibrating on E2 composite success would require impractically broad
dev coverage over subtask compositions, since E2 is closed-loop and
intervention-sensitive; we thus calibrate on E1.

\section{Experiments}\label{sec:experiment}

We instantiate CaB in a first-person Minecraft VLA setting and evaluate
it under the matched deployability discipline using intervention-aware
E1/E2 protocol and a When×How decomposition.

\subsection{Experimental Setting}\label{sec:setting}

\textbf{Environment.} We evaluate CaB in a first-person Minecraft VLA
environment \cite{MineStudio} with RGB observations at 20 Hz and
low-level discrete actions corresponding to human controls
(keyboard/mouse), rather than higher-level action abstractions (e.g.,
functional commands such as craft/place, or options/skill hierarchies
\citep{MineDojo,Voyager,OmniJARVIS,Plan4MC,Optimus2}). To ensure each
instructed task is executable, we fix task-critical initial conditions
and randomize the remaining components, evaluating 50 episodes per task
(distinct randomized seeds). Dev/test sets are seed-disjoint. See
Appendix\textasciitilde{}\ref{sec:environment_detail} for details.

\textbf{Tasks.} Single tasks follow the Jarvis‑VLA protocol, using the
same instruction set/templates to specify goals in natural language
\citep{JarvisVLA}. We evaluate four task groups (craft, combat, mine,
smelt), 8 tasks per group (32 total). Composite tasks are fixed
sequences of two semantically meaningful consecutive tasks drawn from
the same groups (18 composite routines total). Episode horizons are
capped at \(T_{\max}=600 / 1800\) (Single / Composite).

\textbf{E1/E2 protocol and uncertainty.} We follow the
intervention-aware E1/E2 protocol in Sec.\ref{sec:e12} (shared rollout
bank for E1; closed-loop execution for E2). Calibration is dev-only and
frozen on test. We report 95\% bootstrap confidence intervals unless
otherwise stated.

\textbf{Implementation details.} We train on the VPT Demonstration
Dataset \citep{VPT} with a PaliGemma-3B backbone
\citep{PaliGemma,PaliGemma2}; following the \(\pi\)-series robotics VLA
models \citep{pi_zero,pi,pi_Knowledge,pi_six}. Training uses 2× RTX 6000
Ada GPUs for 10 days; we use a single training seed. See
Appendix\textasciitilde{}\ref{sec:implementation_detail} for details.

\subsection{Baselines (matched deployability
discipline)}\label{sec:baseline}

We compare CaB against completion baselines under the matched
deployability discipline (Sec.\ref{sec:discipline}): no test-time
relearning and a single global \((\theta,L)\) calibrated on dev set and
frozen on test set. All methods use the same PaliGemma-3B backbone. As
in many prior VLA/RL designs, baselines use separate heads for
completion/progress and actions, without conditioning actions on
progress.

\begin{itemize}
\tightlist
\item
  \textbf{Binary completion (+dwell):} predicts done/not-done and
  triggers switching via a learning-free dwell rule
  \citep{Classifier-SAFE,Classifier-VLM,Classifier-VLA}.
\item
  \textbf{Hazard-style completion:} estimates near-boundary completion
  likelihood with a hazard model over the past \(L\) steps
  \citep{Hazard}.
\item
  \textbf{Progress regression (STG):} regresses a step-to-event scalar
  (time-to-completion) without modeling post-event behavior
  \citep{STG-SELF,STG-RL,STG-VLM}.
\item
  \textbf{Signed-distance regression:} regresses a signed distance
  \(d_t=\mathrm{clip}(t-t^*,-D,D)\) and converts it to a proximity score
  \(s_t=\max(0,\;K-|d_t|)\). This is a strong event-anchored scalar
  comparator, yet brittle under polarity shifts with a single global
  \((\theta,L)\) (Appendix\textasciitilde{}\ref{sec:polarity}).
\end{itemize}

\textbf{Sanity check:} Under the single-task evaluation protocol, our
models show competitive Success Rate relative to public Minecraft VLA
models (incl.~Jarvis‑VLA Qwen2‑VL‑7B) \citep{JarvisVLA,STEVE1,VPT}. This
comparison is not directly comparable due to different
backbones/training setups (see
Appendix\textasciitilde{}\ref{sec:sanity}).

\subsection{Main Results: Deployable Completion under matched
calibration capacity}\label{sec:main_results}

Table\ref{tab:main_results} summarizes the main results under the
matched deployability discipline (Sec.\ref{sec:discipline}): a single
global \((\theta,L)\) calibrated on dev set and frozen on test set. We
report E1 (action-fixed detection: Completion-F1 / False Completion) and
E2 (closed-loop execution: Single-SR / Composite-SR). To probe
robustness under a single global rule, we additionally report group-wise
E1 Completion-F1 across the task groups and the worst-case E1 regret
\(R_{\text{worst}}(m)=\max_{g\in\mathcal{G}}\big(\max_{m'}\mathrm{F1}_{g,m'}-\mathrm{F1}_{g,m}\big)\),
with \(g\) over task groups and \(m'\) over methods.

\textbf{Key takeaways.} (1) Under the matched deployability discipline,
\textbf{CaB-When} improves E1 detection (Completion-F1\(\uparrow\),
False Completion\(\downarrow\)) and remains robust across task groups,
achieving small worst-case E1 regret (e.g., Signed-distance: 31.9
\(\rightarrow\) 5.8). In contrast, each baseline exhibits at least one
weak task group (large regret), even for the strong event-anchored
signed-distance regressor. (2) \textbf{CaB-When} also improves E2
composite execution (Composite-SR; e.g., 9.1 \(\rightarrow\) 10.4), as
improved completion timing under intervention leads to downstream
success. (3) Adding \textbf{CaB-How} boosts Composite-SR (e.g., 10.4
\(\rightarrow\) 12.6) without materially changing E1, showing that the
\emph{when} interface remains fixed/auditable, while the \emph{how}
consumer reuses the same completion object to stabilize closed-loop
behavior through the handoff.

Overall, consistently low worst-case E1 regret under the matched
discipline supports \textbf{CaB-When} as a \textbf{deployable completion
interface} robust to task-group shifts, while \textbf{CaB-How} further
improves closed-loop composite execution; see
Appendix\textasciitilde{}\ref{sec:qual_bpt_audit} for qualitative
examples.

% Requires: \usepackage{booktabs}

% Table 1: Main scoreboard (template)
\begin{table*}[t]
\centering
\caption{Main scoreboard under the matched deployability discipline (single global $(\theta,L)$; dev-only calibration; frozen on test). E1: action-fixed detection (overall and group-wise F1; worst-case E1 regret). E2: closed-loop execution. Best mean is in bold.}
\label{tab:main_results}
\vspace{1mm}
\small
\setlength{\tabcolsep}{3pt}
\renewcommand{\arraystretch}{0.95}
\begin{tabular}{lcc>{\scriptsize}c ccc}
\toprule
Method & E1: F1$\uparrow$ & E1: FP$\downarrow$ & E1: F1 (Cr/Co/Mi/Sm)$\uparrow$ & E1: $R_{\text{worst}}\downarrow$ & E2: Single$\uparrow$ & E2: Comp$\uparrow$ \\
\midrule
Binary
& $\pmci{73.8}{2.1}$ & $\pmci{36.3}{3.4}$ & $92.2/78.9/41.0/76.5$ & $45.3$ & $\pmci{52.4}{2.1}$ & $\pmci{8.0}{1.9}$ \\
Hazard-style
& $\pmci{77.8}{1.9}$ & $\pmci{30.4}{3.0}$ & $86.5/88.8/46.3/87.7$ & $40.0$ & $\pmci{51.8}{2.1}$ & $\pmci{8.9}{2.0}$ \\
Progress reg
& $\pmci{74.6}{2.2}$ & $\pmci{29.2}{3.3}$ & $55.2/83.2/68.6/89.9$ & $40.3$ & $\pmci{51.4}{2.1}$ & $\pmci{7.4}{1.9}$ \\
Signed-distance reg
& $\pmci{79.5}{2.0}$ & $\pmci{16.5}{2.6}$ & $63.6/88.1/75.7/90.9$ & $31.9$ & $\pmci{52.1}{2.0}$ & $\pmci{9.1}{2.1}$ \\
CaB-When
& $\pmci{90.3}{1.4}$ & $\pmci{13.8}{2.6}$ & $95.5/88.9/80.5/93.2$ & $5.8$ & $\pmci{51.2}{2.1}$ & $\pmci{10.4}{2.2}$ \\
CaB (When+How)
& $\pmcibest{90.5}{1.4}$ & $\pmcibest{13.4}{2.5}$ & $94.1/85.3/86.3/94.2$ & $\best{3.6}$ & $\pmcibest{61.1}{1.9}$ & $\pmcibest{12.6}{2.2}$ \\
\bottomrule
\end{tabular}
\end{table*}

\subsection{When (Switching) × How (Conditioning): A
Decomposition}\label{sec:2x2_results}

To attribute composite gains to \emph{when} the handoff occurs versus
\emph{how} behavior traverses the boundary, we run a 2×2 ablation
varying Switching (FS vs CD) and Conditioning (How-OFF vs How-ON). Here
FS uses a fixed schedule that allocates the episode horizon uniformly
across subtasks, CD uses completion-driven switching with the same
global \((\theta,L)\), and How-ON/OFF enables/disables action
conditioning on the inferred BPT token. (CD corresponds to CaB-When;
How-ON corresponds to enabling CaB-How.)

\textbf{Key takeaways.} Table\ref{tab:2x2_results} reveals a structured
interaction between Switching and Conditioning. \textbf{CD} sharply
reduces overrun (e.g., 92.5 \(\rightarrow\) 4.5) at the cost of a modest
increase in premature switching (e.g., 4.9 \(\rightarrow\) 12.9), and
also improves Composite-SR and \(\mathrm{SR}_{2\mid 1}\) (e.g.,
Composite-SR 6.3 \(\rightarrow\) 10.4; \(\mathrm{SR}_{2\mid 1}\) 8.7
\(\rightarrow\) 15.7), suggesting that better \emph{when} decisions can
benefit downstream behavior by aligning the handoff closer to the
boundary. \textbf{How-ON} improves Composite-SR and
\(\mathrm{SR}_{2\mid 1}\) strongly under FS (e.g., Composite-SR 6.3
\(\rightarrow\) 8.9; \(\mathrm{SR}_{2\mid 1}\) 8.7 \(\rightarrow\) 12.7)
but yields smaller gains under CD (e.g., Composite-SR 10.4
\(\rightarrow\) 12.6; \(\mathrm{SR}_{2\mid 1}\) 15.7 \(\rightarrow\)
18.7), indicating smaller marginal gains once switching is already
boundary-aligned.

Overall, enabling both yields the best Composite-SR (12.6), supporting
complementary contributions from \textbf{CaB-When} (\emph{when}) and
\textbf{CaB-How} (\emph{how}) under the same completion object.

% Requires: \usepackage{booktabs}

% switching_pf_2x2
\begin{table}[t]
\centering
% \caption{When (Switching)$\times$How (Conditioning) 2$\times$2 decomposition (E2): Composite-SR, $\mathrm{SR}_{2\mid 1}$, and boundary diagnostics.}
\caption{When(Switching)$\times$How(Conditioning) decomposition in closed-loop composite execution (E2). FS = fixed-schedule; CD = completion-driven; How-ON/OFF = action conditioning on/off.}
\label{tab:2x2_results}
\resizebox{0.8\columnwidth}{!}{%
\begin{tabular}{ccrrrrr}
\toprule
Switching & How & \multicolumn{1}{c}{E2: Composite-SR $\uparrow$} & \multicolumn{1}{c}{E2: $\mathrm{SR}_{2\mid 1}$ $\uparrow$} & \multicolumn{1}{c}{Premature $\downarrow$} & \multicolumn{1}{c}{Overrun $\downarrow$} \\
\midrule
FS & OFF & $\pmci{6.3}{1.9}$ & $\pmci{8.7}{2.6}$ & $\pmcibest{4.9}{2.1}$ & $\pmci{92.5}{2.5}$   \\
FS & ON  & $\pmci{8.9}{2.1}$ & $\pmci{12.7}{3.0}$ & $\pmci{6.3}{2.3}$ & $\pmci{91.8}{2.6}$  \\
CD & OFF & $\pmci{10.4}{2.2}$ & $\pmci{15.7}{3.3}$ & $\pmci{12.9}{3.2}$ & $\pmcibest{4.5}{2.1}$ \\
CD & ON  & $\pmcibest{12.6}{2.2}$ & $\pmcibest{18.7}{3.0}$ & $\pmci{10.5}{2.8}$ & $\pmcibest{4.5}{2.0}$ \\
\bottomrule
\end{tabular}
}
\end{table}

\section{Analysis and Ablations}\label{sec:ablation}

\textbf{(i) Auditable readout ablations.} Table\ref{tab:ablations}(i)
keeps the trained BPT posterior fixed and varies only the
completion-interface readout (Full / Before-only / After-only / Constant
/ Learnable) under the same dev-only calibration protocol. One-sided or
constant readouts degrade E1 detection, supporting the need to preserve
two-sided boundary evidence, while the dev-only learnable readout yields
only marginal gains over the fixed kernel. See
Appendix\textasciitilde{}\ref{sec:kernel} for details.

\textbf{(ii) Locality via BPT window radius \(K\).}
Table\ref{tab:ablations}(ii) varies the BPT window radius \(K\) under
the same dev-only calibration protocol. Increasing \(K\) leaves E1
detection largely unchanged but reduces E2 Composite-SR, suggesting that
the locality of the BPT window is important for closed-loop execution:
larger windows admit more off-boundary mass and degrade handoffs.

\textbf{(iii) Robustness to noisy completion timestamps.}
Table\ref{tab:ablations}(iii) tests robustness to imperfect offline
completion-time annotation. We perturb training timestamps via
\(\tilde t^* = t^* + \epsilon\), \(\epsilon \sim \mathrm{Unif}[-a,a]\),
sweeping \(a\in\{0, 10, 20\}\), and evaluate with clean \(t^*\). E1/E2
outcomes quantify robustness to train-time timestamp jitter, and E2
remains competitive with baselines trained on clean timestamps.

% Requires: \usepackage{booktabs}

\begin{table*}[t]
\centering
\caption{\textbf{Ablations (i--iii).} (\textbf{i}) Auditable readout variants (E1-only). (\textbf{ii}) BPT window radius $K$ and (\textbf{iii}) train-time timestamp jitter (E1/E2). $(\theta,L)$ is dev-calibrated (E1) and frozen on test.}
\label{tab:ablations}
\vspace{3mm}
\small
\setlength{\tabcolsep}{3pt}
\renewcommand{\arraystretch}{0.95}

% ---------------- (i) ----------------
\textbf{(i) Auditable readout ablation (E1)}\par\vspace{1mm}
\begin{tabular}{lccccccc}
\toprule
Kernel & E1: F1$\uparrow$ & E1: FP$\downarrow$ & E1: F1(Cr)$\uparrow$ & E1: F1(Co)$\uparrow$ & E1: F1(Mi)$\uparrow$ & E1 F1 (Sm)$\uparrow$ & E1: $R_{\text{worst}}\downarrow$ \\
\midrule
Full        & $\pmcibest{90.5}{1.4}$ & $\pmci{13.4}{2.5}$ & $\pmcibest{86.3}{3.8}$ & $\pmci{94.1}{2.0}$ & $\pmcibest{85.3}{3.5}$ & $\pmci{94.2}{2.2}$ & $4.1$ \\
Before-only & $\pmci{78.3}{1.9}$ & $\pmci{46.4}{3.6}$ & $\pmci{63.7}{5.0}$ & $\pmci{82.2}{3.4}$ & $\pmci{72.4}{4.5}$ & $\pmci{95.0}{2.0}$ & $22.6$ \\
After-only  & $\pmci{83.1}{1.8}$ & $\pmcibest{13.1}{2.5}$ & $\pmci{55.5}{7.4}$ & $\pmcibest{98.1}{1.1}$ & $\pmci{78.1}{4.4}$ & $\pmci{83.6}{3.8}$ & $30.8$ \\
Constant    & $\pmci{86.9}{1.6}$ & $\pmci{19.8}{2.9}$ & $\pmci{77.9}{4.5}$ & $\pmci{90.9}{2.5}$ & $\pmci{85.1}{3.6}$ & $\pmci{91.8}{2.6}$ & $8.4$ \\
Learnable   & $\pmcibest{90.5}{1.4}$ & $\pmci{16.9}{2.8}$ & $\pmci{85.5}{4.0}$ & $\pmci{95.9}{1.7}$ & $\pmci{83.3}{3.6}$ & $\pmcibest{95.1}{1.9}$ & $\best{2.3}$ \\
\bottomrule
\end{tabular}

\vspace{2mm} % small separation between (i) and (ii)/(iii)

% ---------------- (ii) and (iii) ----------------
\begin{minipage}[t]{0.49\textwidth}
\centering
\textbf{(ii) BPT window radius $K$ (E1/E2)}\par\vspace{1mm}
\resizebox{\linewidth}{!}{%
\begin{tabular}{ccccc}
\toprule
Radius $K$ & E1: F1$\uparrow$ & E1: FP$\downarrow$ & E2: Comp-SR$\uparrow$ & $\mathrm{SR}_{2\mid 1}\uparrow$ \\
\midrule
$20$ (default) & $\pmci{90.5}{1.4}$ & $\pmci{13.4}{2.5}$ & $\pmci{12.6}{2.2}$ & $\pmci{18.7}{3.0}$ \\
$40$ & $\pmci{87.8}{1.6}$ & $\pmci{14.8}{2.5}$ & $\pmci{8.0}{2.0}$  & $\pmci{12.6}{3.3}$ \\
$60$ & $\pmci{86.9}{1.7}$ & $\pmci{19.0}{2.9}$ & $\pmci{4.8}{1.5}$  & $\pmci{8.0}{2.4}$  \\
\bottomrule

\end{tabular}}
\end{minipage}\hfill% <-- IMPORTANT: no blank line here
\begin{minipage}[t]{0.49\textwidth}
\centering
\textbf{(iii) Timestamp-jitter robustness (E1/E2)}\par\vspace{1mm}
\resizebox{\linewidth}{!}{%
\begin{tabular}{ccccc}
\toprule
Noise level $a$ & E1: F1$\uparrow$ & E1: FP$\downarrow$ & E2: Comp-SR$\uparrow$ & $\mathrm{SR}_{2\mid 1}\uparrow$ \\
\midrule
0 (no noise)      & $\pmci{90.5}{1.4}$ & $\pmci{13.4}{2.5}$ & $\pmci{12.6}{2.2}$ & $\pmci{18.7}{3.0}$ \\
$10$ & $\pmci{87.6}{1.6}$ & $\pmci{25.8}{3.2}$ & $\pmci{8.7}{1.9}$  & $\pmci{12.7}{2.7}$ \\
$20$ & $\pmci{84.3}{1.6}$ & $\pmci{26.5}{3.1}$ & $\pmci{8.5}{1.8}$  & $\pmci{11.6}{2.6}$ \\
% $\epsilon{=}20$ & TBD & TBD & TBD & TBD \\
\bottomrule
\end{tabular}}
\end{minipage}

\end{table*}

\section{Discussion \& Limitations}\label{sec:discussion}

\textbf{Scope and supervision.} CaB uses first-success timestamps
\(t^*\) for supervision during training; these signals can be computed
offline from instrumentation or manual annotation and are not required
at deployment. A limitation is that some tasks may admit ambiguous or
delayed success evidence, in which case proxy or weak event definitions
may be needed, which we leave for future work.

\textbf{Safety, deployment, and societal impacts.} In composites,
completion changes the active instruction, making switching high-impact;
mistiming leaves the agent on the wrong instruction. For safety-critical
deployment, use a deployable completion interface with conservative
calibration, auditable logging; see
Appendix\textasciitilde{}\ref{sec:qual_bpt_audit} for auditability
examples. We also discuss broader impacts in
Appendix\textasciitilde{}\ref{sec:impact}.

\textbf{Conclusion.} Our contribution is not merely predicting
completion, but learning a single boundary-phase posterior that is
dual-used in closed loop---both as a deployable interface for deciding
\emph{when} to switch and as a completion-aware signal for conditioning
\emph{how} actions traverse the handoff.

\bibliographystyle{plainnat}
\bibliography{references}

\clearpage
\appendix
\renewcommand{\thefigure}{\thesection\arabic{figure}}
\renewcommand{\thetable}{\thesection\arabic{table}}
\renewcommand{\theequation}{\thesection\arabic{equation}}
\renewcommand{\theHfigure}{\thesection.\arabic{figure}}
\renewcommand{\theHtable}{\thesection.\arabic{table}}
\renewcommand{\theHequation}{\thesection.\arabic{equation}}
\setcounter{figure}{0}
\setcounter{table}{0}
\setcounter{equation}{0}
\begin{center} {\Large\bfseries Appendix} \end{center}

Appendix Overview. Appendix A discusses broader impacts. Appendices B--G
provide (i) theory and additional analyses/ablations that support the
main claims and (ii) protocol, environment, and implementation details
that support reproducibility. Appendix H provides qualitative audit
examples of the completion interface. Appendix I summarizes assets and
licenses.

\section{Broader Impacts}\label{sec:impact}

\subsection{Positive Impacts}\label{positive-impacts}

CaB targets a practical operational gap in embodied VLA systems:
deciding when an instruction is complete in \emph{closed-loop composite
routines} where switching changes the active instruction and can
materially affect downstream outcomes.

By exposing completion as a deployable completion interface calibrated
once on a dev set and reused unchanged on test, CaB emphasizes
auditable, low-capacity decision rules that may improve reliability
under realistic deployment constraints (e.g., open-ended instruction
spaces where task/group-specific calibration is operationally
ill-defined).

Beyond switching, CaB's completion-aware control conditioning (CaB-How)
is designed to reduce boundary instability during handoffs when
instruction context changes, which may improve predictability in
interactive settings.

\subsection{Negative / Misuse Risks}\label{negative-misuse-risks}

Techniques that improve reliable compositional execution can increase
misuse risk: in broader deployments, they may increase an agent's
ability to faithfully carry out undesirable or harmful composite
instructions if deployed without appropriate instruction-safety
mechanisms and oversight.

More generally, completion-driven switching is a high-impact
intervention: if miscalibrated or triggered under ambiguous success
evidence, the agent may continue operating under an incorrect active
instruction, potentially compounding errors across subsequent subtasks.

While our empirical scope is Minecraft, the underlying interface concept
(deployable completion decisions under limited calibration capacity)
could influence other interactive or embodied settings; this motivates
careful attention to calibration, logging, and monitoring when adapting
the approach beyond controlled simulation.

\subsection{Mitigations and
Safeguards}\label{mitigations-and-safeguards}

We recommend treating completion decisions as an explicitly monitored
interface: since completion changes the active instruction in
composites, switching should be audited as a safety-relevant event in
intervention-sensitive deployments.

Practically, this suggests: (i) conservative calibration of the single
global switching rule on dev data (reused unchanged on test), (ii)
transparent/auditable logging of completion scores and switch triggers
(including false-trigger rates), and (iii) monitoring for systematic
failure modes such as premature or delayed switching under ambiguous
success evidence.

\section{\texorpdfstring{Polarity Shift Analysis: Why Scalar Completion
Signals Are Brittle Under a Single Global Rule
\((\theta,L)\)}{Polarity Shift Analysis: Why Scalar Completion Signals Are Brittle Under a Single Global Rule (\textbackslash theta,L)}}\label{sec:polarity}

This appendix motivates our core design choice: representing the
instruction completion object as an \textbf{event-local posterior over
boundary phase} (Before/Hit/After)---rather than learning completion
only as a single scalar/bit interface signal. In CaB, the posterior is
the object; a scalar is derived only as a low-capacity, auditable
interface readout for switching.

\textbf{Scope (deployability-limited).} We present minimal constructions
illustrating failure modes that arise under our low calibration-capacity
deployability discipline: no test-time relearning, a fixed switching
wrapper, and a single globally reused \((\theta,L)\). These
constructions are not intended as a universal impossibility claim about
all scalar progress/completion signals; rather, they justify why
retaining a boundary-phase posterior is advantageous under this
deployability discipline.

\subsection{Setup and Notation}\label{setup-and-notation}

\textbf{Boundary and boundary-phase variable (event-local signed
distance).}~ For each instruction, completion time is defined by the
first-success event (the \emph{boundary}) at time \(t^*\).~ Let the
signed distance in steps to this boundary be\[
\tilde{\delta}_t := t - t^*.
\]We define the boundary-phase variable as the \emph{event-local clipped
signed distance} \[
\delta_t := \mathrm{clip}(\tilde{\delta}_t,\;[-K,K]) \;=\; \mathrm{clip}(t-t^*,\;[-K,K]),
\] where \(\delta_t<0\) indicates Before, \(\delta_t>0\) indicates
After, and values of \(\delta_t\) near \(0\) correspond to Hit (the
boundary neighborhood). Here \(K\) is the event-local window radius (the
same radius used for BPT windowing and proximity-style readouts).

\textbf{Completion object vs.~scalar/bit interface.}~ A completion
method may expose either

\begin{itemize}
\item
  a \textbf{posterior over boundary phase}, i.e., a posterior over the
  boundary-phase variable \(p(\delta_t\mid c_t)\) (realized in practice
  by discretizing \(\delta_t\) into BPT bins and representing the result
  as a categorical posterior over those bins), which can later be read
  out into a scalar for switching; or
\item
  a \textbf{scalar/bit signal} obtained by collapsing
  \(p(\delta_t\mid c_t)\), e.g., a point estimate
  \(m_t:=\mathbb{E}[\delta_t\mid c_t]\) or a thresholded done bit. Here
  \(m_t\) is the canonical Bayes target for squared-loss regression to
  \(\delta_t\).
\end{itemize}

Here \(c_t\) denotes the past-only context (instruction and
observation/history up to time \(t\)).

\textbf{Deployable switching wrapper (single global \((\theta,L)\)).}
Our deployment discipline uses a fixed, learning-free wrapper: aggregate
a scalar score over a short horizon \(L\) and trigger by a single global
threshold \(\theta\), \[
S_t := \sum_{\tau=t-L+1}^{t} s_\tau,\qquad \text{switch if } S_t \ge \theta,
\] where \(s_t\) is the interface signal, either a scalar/bit or a
posterior readout.

\textbf{Calibration note.} The pair \((\theta,L)\) is calibrated once on
dev and reused unchanged on test (no test-time relearning), matching the
paper's ``fixed wrapper + one global rule \((\theta,L)\)'' protocol.

\textbf{Polarity shift (informal).} Using the boundary-phase variable
\(\delta_t=\mathrm{clip}(t-t^*,[-K,K])\), completion evidence near the
boundary can manifest as posterior mass concentrating on different signs
across tasks: for some tasks, \(p(\delta_t\mid c_t)\) places most mass
on \(\delta_t<0\) (Before-heavy, anticipatory evidence), while for
others it concentrates on \(\delta_t>0\) (After-heavy, confirmatory
evidence). This polarity shift is precisely what challenges scalar/bit
switching under a single global \((\theta,L)\) rule (see Key
Observations 1--2).

\subsection{\texorpdfstring{Key Observation 1: polarity shift +
one-sided supervision \(\Rightarrow\) unreliable under a single global
\((\theta,L)\)
rule}{Key Observation 1: polarity shift + one-sided supervision \textbackslash Rightarrow unreliable under a single global (\textbackslash theta,L) rule}}\label{key-observation-1-polarity-shift-one-sided-supervision-rightarrow-unreliable-under-a-single-global-thetal-rule}

Under polarity shift, tasks differ in which side of the boundary
provides most usable evidence (Before-heavy vs.~After-heavy). Many
progress/step-to-event style baselines supervise a scalar primarily on
one side of the boundary (typically pre-boundary), leaving its behavior
on the other side weakly constrained (often non-identifiable) by the
training objective. Under a single globally reused \((\theta,L)\)
wrapper, a fixed threshold is meaningful only if the scalar behaves
consistently on both sides across tasks; with polarity shift, it is easy
to construct task families whose dominant evidence lies on the
\emph{unsupervised} side, for which the same globally calibrated
threshold becomes unreliable.

This concern is consistent with our main matched calibration results,
where progress-regression baselines (e.g., STG) exhibit brittle
completion switching under the single-global \((\theta,L)\) calibration
(see Table\textasciitilde{}\ref{tab:main_results}).

This motivates either (i) representing completion as a
polarity-invariant proximity scalar derived from a point estimate of
\(\delta_t\), or (ii) representing completion as an explicit
boundary-phase object (Before/Hit/After). Key Observation 2 shows why
option (i) can still be brittle under multi-modal boundary belief,
motivating option (ii).

\subsection{Key Observation 2: Mean-cancellation breaks
proximity-of-mean scalars}\label{sec:proximity-of-mean}

A natural countermeasure to one-sided/phase-sensitive scalars is to use
a polarity-invariant proximity score computed from a point estimate
\(m_t:=\mathbb{E}[\delta_t\mid c_t]\), e.g. \[
s^{\text{prox}}_t := \max(0,\;K-|m_t|),
\] so that ``near the boundary'' corresponds to a large score regardless
of side (Before vs.~After). This removes sign sensitivity, but
introduces a distinct brittleness: collapsing a multi-modal boundary
belief into a single point estimate can yield mean-cancellation,
spuriously appearing ``near boundary.''

\textbf{Why multi-modality is plausible in VLA completion:} near
boundaries, the agent often faces partial observability (visual
occlusions, camera motion, inventory/state not fully visible) and
ambiguous success evidence (weak/latent predicates, delayed effects),
making boundary belief naturally multi-modal even when the policy is
well-trained. In such cases, ``far-before'' vs.~``already-after''
hypotheses can coexist, so a point estimate can be misleadingly close to
zero.

\subsubsection{Proposition (Mean-cancellation false trigger for
proximity of the mean)}\label{prop:proximity-of-mean}

Let \(\delta \in [-K,K]\) denote the boundary-phase variable, i.e., the
event-local clipped signed distance to the first-success event with
window radius \(K>0\). Consider a timestep with a bi-modal posterior
over \(\delta\) given context \(c\): \[
P(\delta=-K\mid c)=\tfrac{1}{2},\qquad P(\delta=+K\mid c)=\tfrac{1}{2},
\] and (for simplicity) zero probability mass for \(|\delta|<K\). Define
the posterior-mean scalar \(m := \mathbb{E}[\delta\mid c]\) and its
proximity transform \[
s_{\text{mean}}:=\max(0,\;K-|m|).
\] Then \(s_{\text{mean}}=K\) (maximal), even though the posterior
places no mass near the boundary.

In contrast, consider the posterior-based proximity readout \[
s_{\text{post}} := \mathbb{E}\!\left[\max(0,\;K-|\delta|)\mid c\right].
\] Then \(s_{\text{post}}=0\), correctly reflecting the absence of
near-boundary mass.

\paragraph{Proof}\label{proof}

By symmetry,
\(m=\mathbb{E}[\delta\mid c]=\tfrac{1}{2}(-K)+\tfrac{1}{2}(+K)=0\).
Hence \(s_{\text{mean}}=\max(0,K-|m|)=K\). But \(\max(0,K-|\delta|)=0\)
whenever \(|\delta|=K\). Therefore
\(s_{\text{post}}=\mathbb{E}[\max(0,K-|\delta|)\mid c]=0\). \(\square\)

\subsubsection{Interpretation}\label{interpretation}

Proposition\textasciitilde{}\ref{prop:proximity-of-mean} shows that even
polarity-invariant proximity scalars computed from a point estimate
(e.g., mean or regressor output) can falsely trigger when boundary
belief is multi-modal: ``far-before OR far-after'' can cancel to
\(m\approx 0\), which proximity-of-mean interprets as ``near boundary.''
This motivates applying proximity to the distribution (posterior mass
near the boundary), not to a single collapsed scalar.

This point-estimate brittleness is consistent with our main-table
results, where the corresponding polarity-invariant proximity baseline
(e.g., Signed-distance reg(+proximity)) remains sensitive under the
single-global \((\theta,L)\) discipline
(Table\textasciitilde{}\ref{tab:main_results}).

\subsection{Connection to CaB-When: Posterior readout as ``proximity of
the
distribution''}\label{connection-to-cab-when-posterior-readout-as-proximity-of-the-distribution}

CaB predicts an event-local posterior \(p_t(y)\) over BPT bins
(including Before/Hit/After around the boundary), a discretized
realization of \(p(\delta_t\mid c_t)\). CaB-When then computes a
transparent scalar score via a fixed linear readout \[
s_t=\sum_{y\in\mathcal{Y}} w_y\,p_t(y),
\] where the weights are predetermined and depend only on bin-center
signed distance (e.g., a triangular proximity kernel
\(w_y\propto \max(0,K-|\delta(y)|)\)).

\textbf{Discrete BPT bridge.} The distribution-level proximity \[
s_{\text{post}}=\mathbb{E}[\max(0,K-|\delta|)\mid c_t]
\] is realized by the discrete BPT posterior as
\(s_t=\sum_y w_y p_t(y)\) with \(w_y=\max(0,K-|\delta(y)|)\).

This is ``proximity of the distribution'' rather than ``proximity of a
point estimate.'' We use this \(s_t\) within the same fixed
\((\theta,L)\) wrapper (dev-calibrated, frozen on test) as in the main
paper.

\subsubsection{Why this matters under limited calibration
capacity}\label{why-this-matters-under-limited-calibration-capacity}

Under a strict single global \((\theta,L)\), robustness depends on
whether the interface score behaves consistently across tasks and
ambiguity patterns. The first observation above highlights a structural
issue of one-sided supervision: post-boundary behavior can be weakly
constrained, making a single frozen threshold unreliable for some task
families. Proposition\textasciitilde{}\ref{prop:proximity-of-mean}
highlights a complementary issue for polarity-invariant point estimates:
multi-modal boundary belief can cause mean-cancellation and false
triggers. By retaining a boundary-phase posterior and applying a fixed
proximity-based readout to posterior mass near the boundary, CaB-When is
designed to remain auditable while being robust to both failure modes
under the paper's deployment discipline.

\section{Evaluation Protocol Details
(E1/E2)}\label{sec:evaluation_detail}

This appendix provides protocol details omitted from the main text due
to space; the main text contains the high-level concept and motivation
in Sec\textasciitilde{}\ref{sec:evaluation}. We specify an
intervention-aware evaluation protocol under matched deployability
constraints, separating (i) action-fixed completion detection on a
shared trajectory bank (E1) from (ii) closed-loop execution with
switching enabled (E2), and we define the metrics, calibration
procedure, and uncertainty estimation used throughout the paper.

\subsection{Overview: E1 vs.~E2 and what is held
fixed}\label{overview-e1-vs.-e2-and-what-is-held-fixed}

\begin{itemize}
\item
  \textbf{E1 (necessary; action-fixed detection).} E1 evaluates
  completion-signal quality on a shared bank of fixed trajectories
  (observations + actions) across all methods, isolating prediction
  quality from behavioral feedback and ensuring matched evaluation
  conditions.
\item
  \textbf{E2 (sufficient; closed-loop execution).} E2 runs the agent
  with switching enabled and measures end-to-end outcomes and boundary
  timing effects under intervention. ~
\end{itemize}

Both E1/E2 follow the same deployability discipline: no test-time
relearning, a fixed low-capacity switching wrapper, and a single global
switching rule calibrated once on dev set.

\subsection{E1 rollout bank
construction}\label{e1-rollout-bank-construction}

We construct a shared set of trajectories that is reused across all
completion methods to ensure matched evaluation conditions in E1,
decoupling completion prediction from policy feedback. ~

\textbf{Equal-mixture bank policy (switching disabled).} To ensure
matched E1 conditions without privileging any single behavior
policy/distribution, we construct the E1 rollout bank using an equal
mixture over the behavior policies corresponding to all baselines and
CaB variants. Concretely, for each episode we sample one behavior policy
uniformly at random from this pool and roll it out in the environment
with the switching wrapper always OFF (i.e., the active instruction is
kept fixed and no completion-driven switching is applied). The resulting
trajectories therefore provide a policy-agnostic and action-fixed
evaluation substrate for E1.

\textbf{What is recorded.} For each episode, we record step-wise:

\begin{itemize}
\tightlist
\item
  RGB observations \(o_t\) and the (fixed) instruction text
  \(l^{(i_t)}\),
\item
  executed actions \(a_t\),
\item
  offline completion annotation \(t^*\) (first-success event) when it
  exists,
\end{itemize}

\textbf{Bank size and splits.} We evaluate 50 episodes per task with
distinct randomized seeds; dev/test splits are seed-disjoint. Dev-only
calibration is performed on the dev bank and frozen for test. ~

\subsection{E1 metrics}\label{e1-metrics}

\textbf{Completion-F1 (Episode-level)}

Let \(t^*\) denote the offline completion time (first-success event) for
an episode when it exists. We use a fixed tolerance window \(\Delta=20\)
steps (1 sec at 20 Hz) around \(t^*\), consistent with the main-text E1
definition. ~

For each episode, we define episode-level indicators:

\begin{itemize}
\tightlist
\item
  \texttt{has\_event}: the episode contains a success event, i.e.,
  \(t^*\) exists.
\item
  \texttt{has\_any\_trigger}: the method triggers completion at least
  once anywhere in the episode.
\item
  \texttt{hit\_in\_window}: the method triggers at least once within the
  tolerance window.
\end{itemize}

Using these indicators, we compute confusion matrix:

\begin{itemize}
\tightlist
\item
  \(TP = \sum \mathbf{1}[\texttt{has\_event} \wedge \texttt{hit\_in\_window}]\)
\item
  \(FN = \sum \mathbf{1}[\texttt{has\_event} \wedge \neg \texttt{hit\_in\_window}]\)
\item
  \(FP = \sum \mathbf{1}[\neg \texttt{has\_event} \wedge \texttt{has\_any\_trigger}]\)
\item
  \(TN = \sum \mathbf{1}[\neg \texttt{has\_event} \wedge \neg \texttt{has\_any\_trigger}]\)
  ~
\end{itemize}

We report Completion-F1: \[
\text{Precision}=\frac{TP}{TP+FP},\quad
\text{Recall}=\frac{TP}{TP+FN},\quad
\text{F1}=\frac{2\,\text{Precision}\,\text{Recall}}{\text{Precision}+\text{Recall}}.
\]

\textbf{False Completion}

We also report False Completion: \[
\text{FalseCompletionRate}=\frac{FP}{FP+TN}.
\] We define False Completion on no-event episodes
(\(\neg\texttt{has\_event}\)); timing errors on event episodes are
characterized in E2 via Premature/Overrun rates. ~

\subsection{E2 metrics}\label{e2-metrics}

\textbf{Task Success Rate}

We report Single Task Success Rate and Composite Task Success Rate under
intervention (switching enabled). Single-task SR is the fraction of
episodes where the target subtask succeeds; Composite-SR is the fraction
of composite episodes where all subtasks succeed within the episode
horizon.

\textbf{Premature Rate / Overrun Rate}

For timing diagnostics, we focus on a sub-instruction with an offline
completion time \(t^*\) (the first-success event). Let \(\hat{t}\) be
the first timestep at which \(\mathrm{Complete}(t)\) fires while this
sub-instruction is active---i.e., the timestep that triggers the switch
to the next sub-instruction. In closed loop, \(\hat{t}\) is the only
meaningful trigger time because the active instruction changes
immediately at \(\hat{t}\).

Using a fixed tolerance window \(\Delta=20\) steps, we define
episode-level timing categories:

\begin{itemize}
\tightlist
\item
  \textbf{Hit-in-window:} \(|\hat{t} - t^*| \le \Delta\).
\item
  \textbf{Premature:} \(\hat{t} < t^* - \Delta\).
\item
  \textbf{Overrun:} \(\hat{t} > t^* + \Delta\), or no trigger occurs
  within the episode horizon.
\end{itemize}

We report Premature Rate and Overrun Rate as episode-level fractions
over evaluation seeds for which \(t^*\) is defined.

\textbf{Handoff quality}

We additionally report handoff quality: \[
\mathrm{SR}_{2\mid 1}=\Pr(\text{subtask 2 succeeds}\mid \text{subtask 1 succeeds}),
\] estimated empirically over composite episodes by conditioning on
episodes where subtask 1 succeeds. ~

\subsection{Calibration protocol}\label{calibration-protocol}

We calibrate the single global rule \((\theta, L)\) once on the dev E1
rollout bank and reuse it unchanged on test (no test-time relearning).
We perform discrete grid search over ranges (rather than task-specific
values): ~

\begin{itemize}
\item
  \textbf{Aggregation horizon \(L\).} We search \(L\) over a small
  discrete set with an upper bound on the order of twice the tolerance
  window, i.e., up to \(\approx 2\Delta\) steps (with \(\Delta=20\)). ~
\item
  \textbf{Threshold \(\theta\).} For each candidate \(L\), we search
  \(\theta\) over the dev min--max range of aggregated scores \(S_t\),
  i.e., \(\theta \in [\min S_t,\; \max S_t]\) computed on the dev bank.
  ~
\end{itemize}

We select the rule \((\theta, L)\) that maximizes episode-level E1
Completion-F1 on dev; ties, if any, are broken by minimizing the
episode-level False Completion rate. The selected \((\theta, L)\) is
then frozen for all test evaluations. ~

\subsection{Uncertainty estimation}\label{uncertainty-estimation}

Unless otherwise stated, we report 95\% bootstrap confidence intervals.
We use a hierarchical bootstrap that preserves the per-task evaluation
budget: for each bootstrap replicate, we resample episodes within each
task with replacement while keeping the number of sampled episodes per
task fixed to the original count (e.g., 50 episodes per task). ~

We then aggregate metrics across tasks (or task groups) in the same
manner as the main report and compute confidence intervals from the
bootstrap distribution (e.g., 2.5/97.5 percentiles). ~Although
percentile bootstrap CIs can be asymmetric, in our experiments the
empirical CIs were nearly symmetric; for readability, we therefore
report them in symmetric ``\(\pm\)'' form using half-widths, i.e., we
display \(\hat{x}\pm \tfrac{1}{2}(U-L)\) while still computing \((L,U)\)
from the bootstrap percentiles.

\section{Minecraft Environment Details}\label{sec:environment_detail}

This appendix specifies the Minecraft evaluation environment used in our
experiments and the interface assumptions under which completion is
studied. We provide concrete definitions of the observation/action
interfaces and task design to support reproducibility and to clarify
what information is (and is not) available to the policy at inference
time.

\subsection{Overview (positioning and interface
assumptions)}\label{overview-positioning-and-interface-assumptions}

Many widely-used Minecraft simulators (e.g., MineDojo
\citep{MineDojo,Voyager}) expose multimodal observations beyond RGB
(e.g., inventory/GPS/compass/voxels) and support functional actions
(e.g., craft/equip/place) that abstract away fine-grained GUI
manipulation; programmatic tasks can further be evaluated via
deterministic success predicates on simulator states. Such simulators
have served as common infrastructure in recent Minecraft agent research;
see, e.g., \citep{OmniJARVIS,Plan4MC,Optimus2,WORLDMEM,MineAnyBuild}.

In contrast, our setting intentionally operates in a \textbf{native
human-interface regime}: the policy consumes only egocentric RGB pixels
(including HUD overlays) and issues low-level keyboard/mouse controls at
a fixed 20 Hz step rate, requiring cursor-based GUI interaction when
applicable. Environment-provided diagnostics may be recorded for
bookkeeping and offline annotation, but are not used as direct policy
inputs. Completion timestamps (first-success events) used for
supervision/evaluation are obtained offline and are not available at
deployment.

We run all evaluations using the MineStudio simulator (v1.1.4)
\citep{MineStudio}, which provides an easily customizable Minecraft
simulator built on MineRL \citep{MineRL} and exposes a low-level
keyboard/mouse control interface.

\subsection{Observation space}\label{observation-space}

\textbf{Raw pixel observations.} Observations are the RGB pixels a human
player would see in first-person Minecraft, including HUD overlays
(e.g., hotbar, health/hunger indicators, and hand animation), rather
than removing them. ~

\textbf{Rendering and model input resolution.} The environment renders
frames at (640, 360) and resizes them to a (224, 224) image before
passing them to the model (rendering resolution for environment
fidelity; model input size for computational efficiency). ~

\textbf{GUI and cursor rendering.} When an in-game GUI is open, the
observation includes a rendered mouse cursor at the current cursor
position, matching the native human-interface visual stream required for
inventory and crafting interactions. ~

\subsection{Action space (Keyboard /
Mouse)}\label{action-space-keyboard-mouse}

Our agent acts through a native human-style interface at 20 Hz, using
discrete keyboard actions together with discretized relative mouse
movements. This interface is designed to support both navigation/combat
and fine-grained GUI interaction within a single action representation.
~

\textbf{Keyboard actions.} The discrete key/button set mirrors standard
Minecraft controls for movement and interaction (20 buttons in total),
including:

\begin{itemize}
\tightlist
\item
  locomotion: move forward / backward / strafe left / strafe right~
\item
  mobility modifiers: jump, sprint, sneak~
\item
  interaction: attack (left-click), use/place (right-click), drop~
\item
  inventory/GUI toggle: open/close inventory~
\item
  hotbar selection: choose one of slots 1--9 (or no hotbar switch) ~
\end{itemize}

\textbf{Mouse movements (relative, discretized).} Mouse movement is
represented as discretized relative X/Y motions: when no GUI is open,
X/Y rotate the first-person camera view (horizontal/vertical look
direction); when a GUI is open, X/Y move the on-screen cursor (which is
rendered in the observation).

To support both coarse camera turns and precise cursor positioning with
a single representation, we discretize mouse movement using foveated
binning along each axis: bins are denser around small motions and
coarser for large motions. We use a 21-bin × 21-bin discretization
(i.e., 441 joint classes for (X,Y) movement). The binning scheme is
otherwise unchanged: each bin corresponds to an interval of relative
motion, and a representative bin center is used when converting a
discrete bin back into a concrete motion at execution time. ~

\subsection{Structured Action Tokenization (Keyboard +
Mouse)}\label{structured-action-tokenization-keyboard-mouse}

A purely factored parameterization (independent Bernoulli decisions per
key and independent mouse-axis outputs) can distort correlated human
control patterns because it cannot represent dependencies among
simultaneous keypresses and mouse movements. At the same time, a single
flat joint distribution over all keys and mouse combinations is
impractical due to combinatorial growth. Following VPT's motivation for
moving beyond fully factored heads \citep{VPT}, we therefore use
factorization to capture dependencies, but we do not introduce an
explicit hierarchical gate.

\textbf{Mutual exclusivity constraints.} Following VPT's action mapping,
we organize mutually exclusive controls into categorical groups (e.g.,
forward vs.~backward; left vs.~right; sprint vs.~sneak; and hotbar
selection), reducing invalid combinations while preserving realistic
human control patterns.

\textbf{Factorized (token-based) modeling.} Consistent with this choice,
we adopt a factorization that is more compatible with Transformer-style
action token sequence modeling. Concretely, we represent each step with
two discrete components---a structured keyboard/action token and a
discretized mouse-movement token---and model their joint distribution
via an ordered factorization: \[
p(a^{\text{key}}_t, a^{\text{mouse}}_t \mid c_t)
= p(a^{\text{key}}_t \mid c_t)\; p(a^{\text{mouse}}_t \mid c_t, a^{\text{key}}_t),
\] where \(c_t\) denotes the past-only context (observation and
instruction history). For CaB-How, we additionally condition on the
phase token by augmenting the context, i.e.,
\(\tilde{c}_t=(c_t, y_t^{\mathrm{cond}})\).

This preserves cross-component dependencies (mouse conditioned on the
keyboard/action decision) without requiring a flat joint over all
combinations. ~

In our implementation, this yields a 4,321-way categorical distribution
for the structured keyboard/action token (after applying
mutual-exclusivity grouping) and a 441-way categorical distribution for
the discretized mouse token (21×21 bins).

\textbf{No-mouse steps.} When no mouse movement is intended, we encode
this as the relative displacement (0, 0), i.e., selecting the
zero-motion (center) bin in the same 21×21 discretization.

\subsection{Tasks, task groups, and composite
construction}\label{tasks-task-groups-and-composite-construction}

\textbf{Task design (Jarvis‑VLA protocol).} We adopt the Jarvis‑VLA
protocol for task design \citep{JarvisVLA}. Concretely, each task is
defined by a natural-language instruction instantiated from a static
instruction set/templates; the same instruction is provided to the agent
at execution time. Task success is evaluated via a
simulator-state--based success predicate, used only for evaluation and
offline annotation (e.g., first-success timestamps), and is never
exposed to the policy at deployment/inference time. Task grouping also
follows the Jarvis‑VLA protocol, as described below.

\textbf{Task groups.} We consider four groups, defined by the dominant
interaction pattern and required environment interface: (i)
\textbf{craft} (crafting/inventory-centric), (ii) \textbf{combat} (enemy
interaction), (iii) \textbf{mine} (resource acquisition), and (iv)
\textbf{smelt} (furnace/processing-centric).

\textbf{Single tasks.} We evaluate a fixed set of 8 single tasks per
group (32 total).

\begin{itemize}
\item
  \textbf{combat (8)}:\\
  zombie, spider, skeleton, sheep, pig, creeper, cow, chicken.
\item
  \textbf{mine (8)}:\\
  oak\_log, stone, coal\_ore, iron\_ore, diamond\_ore, dirt, sand,
  obsidian.
\item
  \textbf{craft (8)}:\\
  bread, crafting\_table, furnace, stick, iron\_pickaxe, iron\_sword,
  diamond\_chestplate, diamond\_boots.
\item
  \textbf{smelt (8)}:\\
  iron\_ingot, gold\_ingot, coal\_from\_smelting, charcoal, glass,
  baked\_potato, cooked\_beef, cooked\_chicken.
\end{itemize}

\textbf{Composite routines.} We evaluate 18 fixed two-step composite
routines formed by selecting semantically meaningful ordered subtask
pairs from a predefined per-group subtask inventory; this inventory
includes the single-task set and may additionally include composite-only
subtasks that do not appear as standalone single tasks. Composite
execution is evaluated under the E2 protocol, where switching is treated
as an intervention, and is not directly supervised as multi-instruction
sequences in the training data.

Each pair is ordered (A precedes B).

\begin{itemize}
\item
  \textbf{mine × craft (5)}:\\
  oak\_log × crafting\_table, oak\_log × stick, stone × furnace, stone ×
  stone\_pickaxe, stone × stone\_sword.
\item
  \textbf{mine × smelt (4)}:\\
  coal\_ore × baked\_potato, oak\_log × charcoal, sand × glass,
  iron\_ore × iron\_ingot.
\item
  \textbf{combat × smelt (4)}:\\
  cow × cooked\_beef, chicken × cooked\_chicken, sheep × cooked\_mutton,
  pig × cooked\_porkchop.
\item
  \textbf{craft × craft (5)}:\\
  oak\_planks × chest, oak\_planks × white\_bed, stick × torch, stick ×
  wooden\_pickaxe, stick × wooden\_shovel.
\end{itemize}

\subsection{Episode semantics: horizons, death, and
respawn}\label{episode-semantics-horizons-death-and-respawn}

\textbf{Episode horizons.} We cap episode length at 600 steps for single
tasks and 1800 steps for composite tasks at 20 Hz, corresponding to 30
sec and 90 sec respectively. ~

\textbf{Death and respawn.} The episode does not terminate on agent
death. Instead, the agent drops its items and respawns in the initial
spawn region and continues the same episode. This
non-termination-on-death design keeps the evaluation budget comparable
across episodes and avoids confounding completion timing with early
termination.

\section{Implementation Details}\label{sec:implementation_detail}

This appendix summarizes implementation choices held fixed across all
methods to ensure matched comparisons. We specify the I/O interfaces
(action and completion tokenization), then describe backbone adaptation,
data preprocessing, training objectives, distributed training, and
compute used in our experiments.

\subsection{Fixed interfaces \&
factorizations}\label{fixed-interfaces-factorizations}

We fix the discrete output interfaces used throughout the paper. Control
is represented by two action tokens: a structured button/action token
and a discretized mouse-movement token. The button/action token has
vocabulary size \(|V_{\mathrm{btn}}|=4321\) (after applying
mutual-exclusivity grouping), and the mouse token uses a \(21\times 21\)
discretization with \(|V_{\mathrm{mouse}}|=441\). Completion is
represented by a Boundary-Phase Token (BPT) with vocabulary size
\(|\mathcal{Y}|=20\), whose binning (including out-of-window handling)
is specified in Appendix.\ref{sec:training_conf}.

At each decision step \(t\), we generate completion and control in an
ordered manner. Let \(c_t\) denote the past-only context
(observation/instruction history), \(y_t\in\mathcal{Y}\) the BPT token,
and \((a^{\mathrm{btn}}_t, a^{\mathrm{mouse}}_t)\) the action tokens. We
model the joint distribution with the ordered factorization

\[
p_\phi\!\left(y_t, a^{\mathrm{btn}}_t, a^{\mathrm{mouse}}_t \mid c_t\right)
=
p_\phi\!\left(y_t \mid c_t\right)\;
p_\phi\!\left(a^{\mathrm{btn}}_t \mid c_t, y_t\right)\;
p_\phi\!\left(a^{\mathrm{mouse}}_t \mid c_t, y_t, a^{\mathrm{btn}}_t\right),
\]

which is compatible with the BPT→action construction used by CaB-How
(Sec.\ref{sec:cab_how}) while making explicit the two-token action
interface used in our implementation.

\subsection{Backbone (PaliGemma) \& post-training
recipe}\label{sec:backbone}

\textbf{Shared backbone and matched I/O.} All methods (CaB and
baselines) are instantiated on the same pretrained backbone and share
the same video-conditioned input interface and chunked action output
interface, so differences arise only from the completion/control
components described in the Baselines section.

\textbf{Backbone initialization and video-context adaptation.} We
initialize from the publicly available image-VLM checkpoint
\textbf{paligemma-3b-pt-224} and adapt it to video-conditioned control
\citep{PaliGemma,PaliGemma2}. While the released PaliGemma
weights/implementation target single-image inputs, our policies consume
a short history of past RGB observations as context. We implement this
as frame stacking with \(F=4\) most recent frames at 20 Hz (a 0.2-second
visual context window). While our implementation supports configurable
frame sampling (e.g., skipping frames), all reported results use
contiguous frames (no skipping).

\textbf{Action chunking and inference stride.} Both CaB and baselines
predict an action chunk rather than a single-step action. During
training we use chunk length \(C=6\) (next 6 low-level actions). At
inference we run the policy every 4 steps and output a chunk of \(C=4\)
low-level actions, executed open-loop until the next policy call; the
chunk length is implementation-configurable. We report these settings
for all main results.

\textbf{What is trained vs.~frozen.} To preserve the pretrained visual
representation while adapting to control, we freeze the PaliGemma vision
encoder and train the remaining components: the Transformer, the
vision-to-token projector, and the action head(s) (including
completion-related heads required by CaB). This training policy is
shared across methods for matched comparisons.

\textbf{Separated vocabularies and heads (text vs.~action).} We
explicitly separate the text-token vocabulary/head used by the
pretrained VLM from the action-token vocabulary/head introduced for VLA
control. Action-loss gradients update the action pathway and shared
backbone, without contaminating the pretrained text head. It also avoids
common VLA engineering shortcuts that repurpose a small set of reserved
tokens or low-frequency text tokens as action symbols.

\textbf{How-ON/OFF implementation (masking-level ablation).} How-ON uses
the standard causal attention mask (allowing BPT→action conditioning as
described in the Method section). How-OFF is implemented by modifying
the attention mask so that action tokens are prevented from attending to
the BPT token, while keeping the rest of the architecture, heads, and
supervision unchanged; this isolates the effect of control-side
conditioning.

\subsection{Data \& preprocessing}\label{sec:training_data}

\textbf{Training data.} We post-train on the \textbf{VPT Minecraft
Demonstration Dataset} \citep{VPT}, which consists of long human
demonstration rollouts captured with a dedicated Minecraft recording
pipeline. Each recording provides a synchronized first-person video
stream and per-step action logs: frames are written every game tick by
downsampling the window framebuffer to 640×360, while actions are logged
separately in a step-aligned format. The recording setup fixes key
rendering settings and captures native human controls (keyboard presses
and continuous mouse movements), rather than relying on a simplified
functional action interface.

\textbf{Why event-centered single-task crops.} Our supervision is
derived from event timestamps: we can identify when a relevant success
event occurs but do not have composite routines annotated with
multi-instruction supervision in the dataset. When an event timestamp is
available, we associate it with the corresponding instruction and
construct a single-task training example around that event. Composite
execution is not directly supervised as multi-instruction sequences in
the training data; instead, it is evaluated at test time via
completion-driven switching under intervention (E2).

\textbf{Asymmetric cropping around completion events.} We construct
training examples by randomly cropping short segments from long rollouts
around the offline completion event time \(t^*\) (first success). The
crop window is intentionally asymmetric: we allow crop starts from far
before the event (up to 200 steps pre-event) to capture anticipatory
context, while limiting the post-event range to within \(K\) plus a
small buffer, where \(K\) is the BPT window radius used in label
construction. This asymmetry reflects the event-timestamped nature of
the data (event times are known, but full composite-instruction
structure is not) and keeps supervision focused on boundary-relevant
behavior.

\textbf{Video context / Chunked targets.} We use the same frame-stacking
and action-chunking specifications as in
Sec.\textasciitilde{}\ref{sec:backbone} for all reported results.

\textbf{Optional expert synthesis for GUI-centric tasks.} For structured
GUI-based tasks (e.g., crafting and smelting), we synthesize a small set
of expert demonstration entries that emphasize inventory/crafting
interactions to better cover interface-heavy behaviors. These
synthesized entries are used only for training and are excluded from
dev/test evaluation banks to preserve matched comparisons.

\subsection{Training specification: objectives, labels, and
optimization}\label{sec:training_conf}

\textbf{Objective.} We train a single autoregressive model to predict
(i) a Boundary-Phase Token (BPT) and (ii) an action output, using
teacher forcing. At each step \(t\), the training loss is the sum of
next-token cross-entropies: \[
\mathcal{L} \;=\; \sum_t \Big(
\mathrm{CE}(y_t,\; p_\phi(y_t \mid c_t))
\;+\;
\mathrm{CE}(a_t,\; p_\phi(a_t \mid c_t, y_t^{\mathrm{cond}}))
\Big),
\] where \(y_t\) is the event-derived BPT label and
\(y_t^{\mathrm{cond}}\) is the token fed to the action head
(ground-truth BPT under teacher forcing). How-ON/OFF differs only in
whether action prediction is allowed to attend to the BPT token (How-ON)
or is prevented from attending to it via masking (How-OFF), while
keeping the same BPT prediction and supervision unchanged.

\textbf{BPT label construction and binning.} Offline completion times
\(t^*\) (first-success events) are used to construct BPT supervision
during training/evaluation but are not available at deployment. We
define signed distance to the boundary \(d_t = t - t^*\) and assign a
discrete BPT label \(y_t \in \mathcal{Y}\) within an event-local window
with radius parameter \(K\) (default \(K=20\)). We use a \(\Delta=2\)
step binning with an expanded hit region: we set \(y_t=\mathrm{Hit}\)
for \(|d_t|\le 1\). For the remaining in-window distances, i.e.,
\(2\le |d_t|\le K-1\), we bin by magnitude using
\(b=\left\lfloor |d_t|/2\right\rfloor\) and assign
\(\mathrm{Before}[b]\) if \(d_t<0\) or \(\mathrm{After}[b]\) if
\(d_t>0\). For \(|d_t|\ge K\), we assign an explicit out-of-window class
\(\varnothing_{\mathrm{ow}}\), which is included in the training loss.

Under this definition, bins correspond to
\(\mathrm{After}[b]\leftrightarrow \{2b,2b+1\}\) and
\(\mathrm{Before}[b]\leftrightarrow \{-(2b+1),-2b\}\). With this scheme,
\(|\mathcal{Y}| = 2\lfloor (K-1)/2\rfloor + 2 = 20\) (Before/After bins
+ Hit + \(\varnothing_{\mathrm{ow}}\)). Whenever a representative signed
distance is required, we use the bin center
\(\delta(\mathrm{After}[b])=2b\), \(\delta(\mathrm{Before}[b])=-2b\),
\(\delta(\mathrm{Hit})=0\); for convenience we set
\(|\delta(\varnothing_{\mathrm{ow}})|=K\) (used only inside
proximity-style weights).

\textbf{Optimization hyperparameters (numbers-only).} We use AdamW with
a cosine learning-rate schedule with warmup:

\begin{itemize}
\tightlist
\item
  Learning rate: \(2\times 10^{-5}\)
\item
  Weight decay: \(0.1\)
\item
  Max grad norm: \(1.0\)
\item
  Warmup ratio: \(0.03\) (cosine schedule with warmup)
\item
  Global batch size: \(128\) (across all GPUs and gradient accumulation)
\item
  Context length: \(1024\)
\item
  Precision: bfloat16
\end{itemize}

\subsection{Systems and resource-aware
training}\label{sec:distributed_training}

\textbf{Distributed training stack.} We build on the Prismatic VLMs
codebase \citep{Prismatic} (also used by OpenVLA \citep{OpenVLA}) and
extend it to support a PaliGemma backbone under PyTorch FSDP2
\citep{FSDP}. We shard transformer blocks and write consolidated
checkpoints (trainable modules by default).

\textbf{Optimizer choices under FSDP2 (resource-aware design).} A
practical bottleneck for VLA post-training is optimizer-state memory:
AdamW maintains first and second-moment states, which increases memory
usage. To improve accessibility, we support memory-efficient optimizers
as drop-in options under the same FSDP2 training loop, including
Adafactor (factored second-moment statistics) \citep{Adafactor} and
APOLLO-style optimizers \citep{APOLLO} designed to reduce
optimizer-state memory.

This enables extensive ablations in a comparatively low-resource setting
(2× RTX 6000 Ada), and we plan to release code enabling these
distributed optimizer choices.

\subsection{Compute / runtime}\label{sec:compute}

\textbf{Training compute.} Unless otherwise stated, post-training
experiments were run on 2\(\times\) NVIDIA RTX 6000 Ada GPUs. We use a
single training seed.

Peak GPU memory during training was 48 GB per GPU, and peak memory at
inference was 13 GB. Training a single model in the main setting took 10
days of wall-clock time on 2× RTX 6000 Ada GPUs (480 GPU-hours).

\textbf{Evaluation compute.} E1 evaluates completion signals on a fixed
rollout bank and consists of offline forward passes over recorded
trajectories, while E2 runs closed-loop rollouts with switching enabled.

For a single trained model, one full pass of the E2 evaluation suite
took approximately 1 day of wall-clock time on the same 2× RTX 6000 Ada
setup.

We report these compute and memory figures to clarify the resource
requirements under which the reported results and ablations were
obtained.

\section{Readout-kernel Analysis}\label{sec:kernel}

\subsection{Auditable readout ablations (completion-interface
readout)}\label{auditable-readout-ablations-completion-interface-readout}

\textbf{Goal and controlled factors.} This ablation isolates the
\emph{completion-interface readout} while holding the learned completion
signal fixed. We keep the trained CaB model unchanged (CaB(When+How))
and reuse its inferred Boundary-Phase Token (BPT) posterior
\(p_t(y)=P_\phi(y_t=y\mid c_t)\) at every step. We vary \emph{only} the
readout that maps the BPT posterior to a scalar completion-evidence
score, and evaluate all variants under the same dev-only calibration
discipline (single global rule calibrated once on dev and reused
unchanged on test; no test-time relearning).

\subsection{Common score form (fixed
kernels)}\label{common-score-form-fixed-kernels}

For fixed, auditable readouts, we compute a scalar score by a
transparent linear functional of the posterior: \[
s_t \;=\; \sum_{y\in\mathcal{Y}} w_y\, p_t(y),
\] where \(\mathcal{Y}\) is the shared BPT vocabulary and \(w_y\) are
fixed weights that do not depend on the task or instruction.

\subsection{Fixed kernels: Full / Before-only / After-only /
Constant}\label{fixed-kernels-full-before-only-after-only-constant}

Let \(\delta(y)\) denote the representative signed distance associated
with BPT class \(y\) (e.g., bin-center distance relative to the
first-success boundary), and let \(K\) be the BPT window radius in
steps. We define the \textbf{full} triangular kernel, the
\textbf{one-sided} kernels, and the \textbf{constant (flattened)} kernel
as: \[
\begin{aligned}
w^{\mathrm{full}}_y   &= \max(0,\,K-|\delta(y)|),\\
w^{\mathrm{before}}_y &= \max(0,\,K-|\delta(y)|)\cdot\mathbf{1}[\delta(y)\le 0],\\
w^{\mathrm{after}}_y  &= \max(0,\,K-|\delta(y)|)\cdot\mathbf{1}[\delta(y)\ge 0],\\
w^{\mathrm{const}}_y  &= \mathbf{1}[y\neq \text{ow}].
\end{aligned}
\] The Full kernel preserves two-sided boundary evidence (Before and
After), the one-sided variants ablate one side by construction, and the
Constant kernel discards signed phase structure while retaining only
total in-window mass
\(s_t=\sum_{y\neq\text{ow}}p_t(y)=1-p_t(\text{ow})\).

\subsection{Learnable readout (dev-only;
higher-capacity)}\label{learnable-readout-dev-only-higher-capacity}

For the \textbf{learnable} readout, we fit a one-hidden-layer MLP that
takes the BPT posterior vector \(p_t\in\mathbb{R}^{|\mathcal{Y}|}\) as
input and outputs a scalar completion score: \[
s_t \;=\; \sigma\!\left(\mathbf{w}_2^\top \,\mathrm{ReLU}(\mathbf{W}_1 p_t + \mathbf{b}_1) + b_2\right),
\] where \(\mathrm{ReLU}(\cdot)\) is the hidden-layer nonlinearity and
\(\sigma(\cdot)\) is the sigmoid.

\textbf{Labels.} We construct binary targets from the E1 tolerance
window around the ground-truth completion time \(t^*\): frames within
the window are labeled positive, and frames outside are labeled
negative.

The MLP is trained on the dev set only; evaluation uses the same fixed
switching wrapper and dev-only calibration protocol as CaB-When (single
global calibration on dev, reused unchanged on test; no test-time
relearning).

\subsection{Results and brief
discussion}\label{results-and-brief-discussion}

All readout variants are evaluated with the same E1 detection metrics
reported in the main text (Completion-F1 / False Completion), and
results are summarized in Table\textasciitilde{}\ref{tab:ablations}(i).

Overall, these readout ablations are complementary to
Appendix\textasciitilde{}\ref{sec:polarity} under the same single-global
\((\theta,L)\) discipline. Holding the BPT posterior fixed, the
completion interface is consumed by a single globally reused
\((\theta,L)\) wrapper, so the readout must be robust across tasks whose
boundary evidence is polarity-shifted (Before-heavy vs.~After-heavy
evidence). One-sided readouts (Before-only / After-only) discard
evidence from one side at the interface level, which can make a single
globally calibrated threshold less reliable across such task variation.
By contrast, the full (two-sided) kernel preserves boundary-phase
structure in the interface score and improves discrimination between
``near-but-before'' and ``near-but-after'' states under the same
calibration capacity.

Finally, the small gap between the dev-only learnable MLP and the fixed
full kernel suggests that, under the single-global \((\theta,L)\)
deployability discipline, robustness is driven less by readout capacity
than by preserving distributional, two-sided boundary evidence in the
completion object. In other words, once the BPT posterior retains both
Before/After information, a simple auditable proximity-style kernel
already extracts most of what the fixed wrapper needs, while
higher-capacity readouts offer limited additional benefit when
calibration capacity is held constant.

\section{Sanity check: CaB vs.~public Minecraft VLA
baselines}\label{sec:sanity}

We additionally report an external-reference sanity check on single-task
success rate (Single-SR) against public Minecraft VLA baselines,
including JARVIS‑VLA (Qwen2‑VL‑7B) and other public baselines (e.g.,
VPT, STEVE‑1) \citep{JarvisVLA,STEVE1,VPT}.

This sanity check is intended to contextualize the overall strength of
our experimental pipeline. It is not an apples-to-apples comparison with
our main matched-discipline results, since the referenced systems differ
in backbone size, training data/objectives, and inference setups.

\textbf{Protocol alignment.} We report Single-SR under our E2
single-task evaluation, which is designed with reference to the
single-task protocol used in prior work \citep{JarvisVLA}. Results are
summarized in Table\textasciitilde{}\ref{tab:sanity_single_sr}. Our goal
is not to claim a matched advantage over external baselines, but to
verify that our models operate in a reasonable performance regime under
an established protocol.

\textbf{Note on VPT.} VPT \citep{VPT} is included as a widely used
demonstration-trained reference point (behavioral cloning on human
gameplay demonstrations). We use it to anchor the scale of Single-SR
under the same measurement procedure and to confirm that the evaluation
reflects instruction-following performance beyond imitation-only
baselines.

\textbf{Summary.} As shown in
Table\textasciitilde{}\ref{tab:sanity_single_sr}, our models achieve
competitive Single-SR relative to the public reference points, and CaB
further improves Single-SR within our pipeline. This sanity check
complements the paper's main focus---deployable completion under
intervention---whose primary claims are established under the matched
deployability discipline in
Table\textasciitilde{}\ref{tab:main_results}.

% Requires: \usepackage{booktabs}

% switching_pf_2x2
\begin{table}[t]
\centering
\caption{External single-task sanity check (same evaluation protocol).}
\label{tab:sanity_single_sr}
\small
\setlength{\tabcolsep}{4pt}
\renewcommand{\arraystretch}{0.95}
\begin{tabular}{lc}
\toprule
Model & Single-SR$\uparrow$ \\
\midrule
VPT & $\pmci{4.2}{0.9}$ \\
STEVE-1 & $\pmci{19.7}{1.6}$ \\
JARVIS-VLA (Qwen2-VL-7B) & $\pmci{45.5}{2.1}$ \\
Baseline (Signed-distance reg) & $\pmci{52.1}{2.0}$ \\
CaB(When+How) & $\pmci{61.1}{1.9}$ \\
\bottomrule
\end{tabular}
\end{table}

\section{Qualitative Audit of the Completion Interface (BPT
Posterior)}\label{sec:qual_bpt_audit}

\begin{figure*}[!t]
  \centering
  \includegraphics[width=\textwidth, height=0.85\textheight, keepaspectratio]{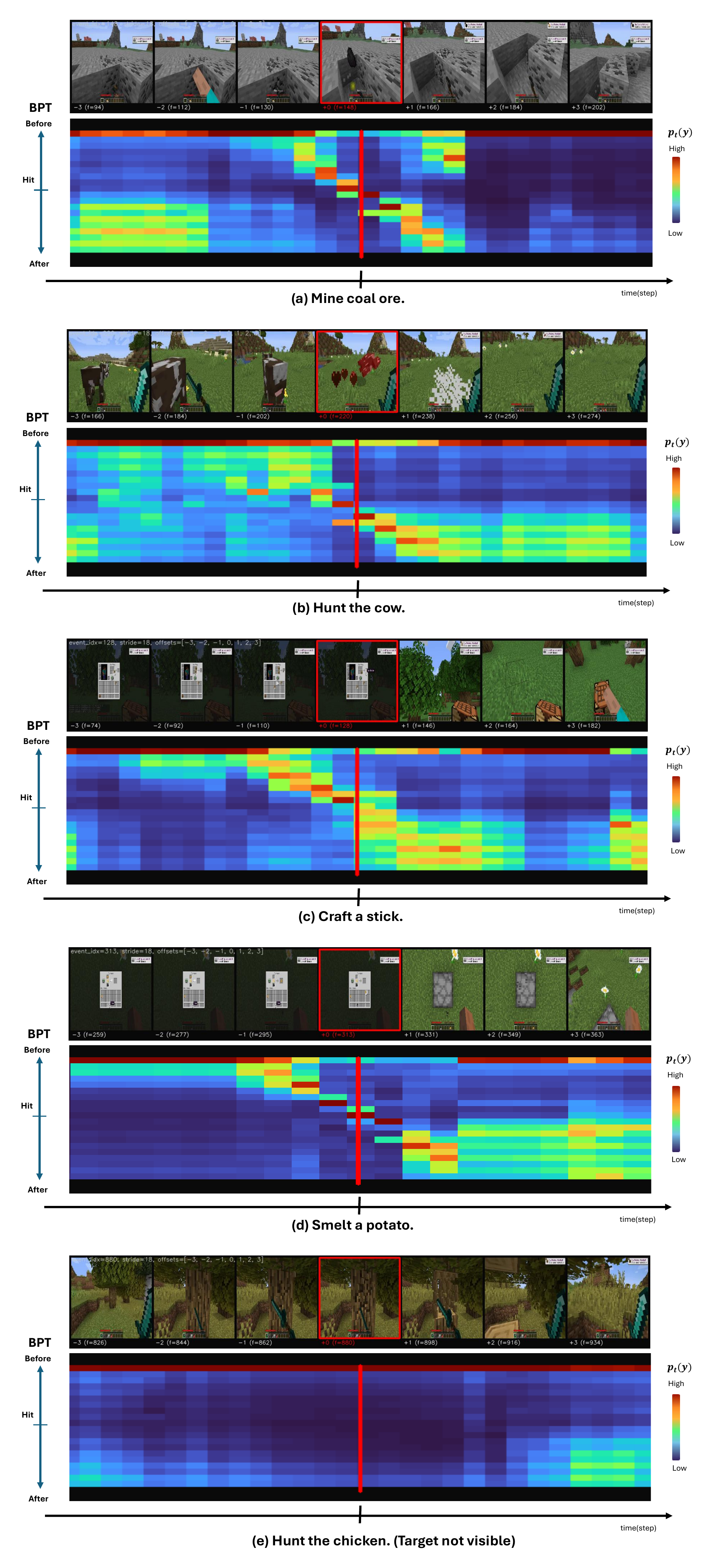}
\caption{\textbf{Qualitative audit of the completion interface.}
We plot the predicted BPT posterior $p_t(y)$ over boundary phase (Before/Hit/After) versus time (steps) during execution. Panels (a--d) show successful episodes spanning mine/combat/craft/smelt, while panel (e) is a \emph{target-not-visible} example (Hunt the chicken) where the posterior correspondingly does not localize near Hit.}
\label{fig:bpt_analysis}
\end{figure*}

This section provides qualitative examples that complement the
quantitative results by visualizing how the predicted Boundary-Phase
Token (BPT) posterior evolves over time during execution.
Figure\textasciitilde{}\ref{fig:bpt_analysis} plots the posterior
\(p_t(y)\) over boundary phase (Before / Hit / After) versus time
(steps) for representative single-instruction episodes: four successful
runs (a--d) spanning mine/combat/craft/smelt, and one target-not-visible
example (e).

\textbf{What is plotted.} For each episode, we plot the model's
predicted BPT posterior \(p_t(y)\) over the boundary-phase classes
(Before/Hit/After) as the agent executes actions under a fixed
instruction (switching disabled). This posterior is the same completion
object that would be read out for switching.

\textbf{Successful episodes: a common qualitative signature (a--d).}
Across four successful tasks---Mine coal ore, Hunt the cow, Craft a
stick, and Smelt a potato---the posterior typically exhibits a
phase-localization pattern: it is Before-heavy earlier in the episode,
then concentrates near Hit around completion, and subsequently shifts
toward After. While the sharpness and persistence of Hit/After evidence
vary across task types, a shared qualitative signature is the emergence
of a coherent boundary-phase transition in \(p_t(y)\).

\textbf{Target-not-visible example: absence of boundary evidence (e).}
In panel (e) (Hunt the chicken; Target not visible), the observation
stream lacks the target entity, so the agent receives little to no
boundary-relevant evidence for completion. Accordingly, the BPT
posterior does not form a clear Hit concentration and remains dominated
by non-hit phase mass throughout the episode. This example highlights
that the completion interface behaves conservatively when task evidence
is missing, rather than spuriously ``firing'' without visual support.

\textbf{Auditability takeaway.} These examples illustrate why the
proposed completion interface is \emph{auditable}: the interface exposes
a structured, time-resolved evidence trace \(p_t(y)\) over interpretable
boundary phases (Before/Hit/After) that a user can directly inspect to
understand \emph{why} a trigger would (or would not) occur. In
successful runs, the posterior localizes around Hit near completion; in
the target-not-visible run, the absence of Hit localization provides an
explicit, human-verifiable indication that completion evidence is not
present.

\section{Assets and Licenses}\label{sec:license}

\subsection{Environments and
Simulators}\label{environments-and-simulators}

\begin{itemize}
\tightlist
\item
  \textbf{Minecraft simulator/tooling:} MineStudio.

  \begin{itemize}
  \item
    \textbf{Version:} v1.1.4 (release).
  \item
    \textbf{URL:}

    \begin{itemize}
    \tightlist
    \item
      GitHub: https://github.com/CraftJarvis/MineStudio
    \item
      Docs: https://craftjarvis.github.io/MineStudio/
    \item
      Release notes: https://github.com/CraftJarvis/MineStudio/releases
    \end{itemize}
  \item
    \textbf{License / Terms:} MIT License (repository license).
  \item
    \textbf{Usage note:} See Appendix.\ref{sec:environment_detail}.
  \end{itemize}
\end{itemize}

\subsection{Datasets}\label{datasets}

\begin{itemize}
\tightlist
\item
  \textbf{Training dataset:} OpenAI VPT contractor demonstrations.

  \begin{itemize}
  \item
    \textbf{Source / access:}\\
    The VPT project distributes code and associated resources via the
    official repository:\\
    https://github.com/openai/Video-Pre-Training
  \item
    \textbf{Primary reference:}\\
    The VPT paper states that the contractor data is open-sourced as
    part of the project's released assets.~

    \begin{itemize}
    \tightlist
    \item
      Paper (OpenAI PDF): https://cdn.openai.com/vpt/Paper.pdf~~
    \item
      NeurIPS proceedings PDF:
      https://proceedings.neurips.cc/paper\_files/paper/2022/file/9c7008aff45b5d8f0973b23e1a22ada0-Paper-Conference.pdf~~
    \end{itemize}
  \item
    \textbf{Version / subset:}\\
    We use a mixture of contractor demonstration subsets (covering
    multiple recorder ``series'' such as free gameplay / early game /
    house building / obtain-diamond-style trajectories) as provided by
    the VPT release.
  \item
    \textbf{License / Terms (dataset):}\\
    The VPT paper indicates the contractor data is open-sourced, but it
    does not specify a standard dataset license in the paper text. We
    therefore do not redistribute the dataset in our release.
  \item
    \textbf{License / Terms (VPT repository):} The VPT repository itself
    is MIT-licensed.
  \item
    \textbf{Usage note:} See Appendix.\ref{sec:training_data}.
  \end{itemize}
\end{itemize}

\subsection{Pretrained Models /
Backbones}\label{pretrained-models-backbones}

\begin{itemize}
\item
  \textbf{Backbone model:} PaliGemma‑3B (VLA backbone).
\item
  \textbf{Checkpoint ID:} google/paligemma-3b-pt-224.~
\item
  \textbf{URL / access:}
  https://huggingface.co/google/paligemma-3b-pt-224 (license-gated
  access).
\item
  \textbf{License / Terms:} Gemma license / Google usage terms
  (acceptance required to access files).
\item
  \textbf{Usage note:} See Appendix.\ref{sec:backbone}.
\end{itemize}

\subsection{External Codebases and
Libraries}\label{external-codebases-and-libraries}

\begin{itemize}
\tightlist
\item
  \textbf{Training codebase foundation:} Prismatic VLMs (MIT).

  \begin{itemize}
  \item
    \textbf{Version (pinned commit):}
    874c5bbff52b248294a3ab97006491a7faa698e6
  \item
    \textbf{URL:}\\
    https://github.com/TRI-ML/prismatic-vlms~
  \item
    \textbf{Pinned commit URL:}\\
    https://github.com/TRI-ML/prismatic-vlms/commit/874c5bbff52b248294a3ab97006491a7faa698e6~
  \item
    \textbf{License:} MIT License (repository LICENSE).
  \item
    \textbf{Usage note:} See Appendix.\ref{sec:distributed_training}.
  \end{itemize}
\item
  \textbf{Core framework:} PyTorch (distributed training with FSDP2 /
  \texttt{fully\_shard}).

  \begin{itemize}
  \item
    \textbf{Version:} 2.7.0 (as used in experiments).
  \item
    \textbf{URL:}~

    \begin{itemize}
    \tightlist
    \item
      Repository: https://github.com/pytorch/pytorch
    \item
      FSDP2 API docs (\texttt{fully\_shard}):
      https://docs.pytorch.org/docs/stable/distributed.fsdp.fully\_shard.html
    \end{itemize}
  \item
    \textbf{License / Terms:} PyTorch is distributed under a BSD‑3
    license (see PyTorch forums clarification).
  \end{itemize}
\end{itemize}

\end{document}